\documentclass[10pt,twocolumn,letterpaper]{article}

\usepackage[pagenumbers]{wacv} % To force page numbers, e.g. for an arXiv version

\usepackage{graphicx}
\usepackage{amsmath}
\usepackage{amssymb}
\usepackage{booktabs}
\usepackage{url}
\usepackage[accsupp]{axessibility} % Improves PDF readability for those with disabilities.

\usepackage[utf8]{inputenc} % allow utf-8 input
\usepackage[T1]{fontenc}    % use 8-bit T1 fonts
\usepackage{url}            % simple URL typesetting
\usepackage{booktabs}       % professional-quality tables
\usepackage{amsfonts}       % blackboard math symbols
\usepackage{nicefrac}       % compact symbols for 1/2, etc.
\usepackage{microtype}      % microtypography
\usepackage{xcolor}         % colors

\usepackage{graphicx}
\usepackage{algpseudocode}
\usepackage{amsmath}
\usepackage{amssymb}
\usepackage{mathtools}
\usepackage{amsthm}
\usepackage{color, colortbl}
\usepackage{algorithm}
\usepackage{algpseudocode}
\usepackage{varwidth}
\definecolor{LightCyan}{rgb}{0.88,1,1}
\definecolor{LightRed}{rgb}{1,0.88,0.88}
\definecolor{Gray}{rgb}{0.8,0.8,0.8}
\definecolor{LightGray}{rgb}{0.92,0.92,0.92}
\definecolor{LightPurple}{RGB}{226, 225, 254}
\usepackage{multirow}
\usepackage{xcolor}
\usepackage{caption}
\usepackage{comment}
\usepackage{wrapfig}
\usepackage{titletoc}
\usepackage{diagbox}
\usepackage{graphicx}
\usepackage{subcaption}

\renewcommand{\P}{\mathbb{P}}

\newcommand{\Ax}{\mathbb{A}}

\newcommand{\E}{\mathbb{E}}

\newcommand{\blue}[1]{\textcolor{black}{#1}}

\newtheorem{theorem}{Theorem}
\newtheorem{lemma}[theorem]{Lemma}
\newtheorem{definition}[theorem]{Definition}
\newtheorem{remark}[theorem]{Remark}
\newtheorem{fact}[theorem]{Fact}

\newtheorem*{theorem*}{Lemma}

\usepackage[pagebackref,breaklinks,colorlinks]{hyperref}
\usepackage{tikz}

\usepackage[capitalize]{cleveref}
\crefname{section}{Sec.}{Secs.}
\Crefname{section}{Section}{Sections}
\Crefname{table}{Table}{Tables}
\crefname{table}{Tab.}{Tabs.}
\def\wacvPaperID{90} % *** Enter the WACV Paper ID here
\def\confName{WACV}
\def\confYear{2025}

\begin{document}

\title{CRAFT: Cross-modal Aligned Features Improve Robustness of Prompt Tuning}

% Cross Alignment Feature Tuening

\author{
Jingchen Sun\thanks{Corresponding authors: Jingchen Sun and Vishnu Suresh Lokhande}\hspace{0.5cm}Rohan Sharma\hspace{0.5cm}Vishnu Suresh Lokhande\footnotemark[1]\hspace{0.5cm}Changyou Chen \\
% {\tt\small \{jsun39, rohanjag, vishnulo, changyou\}@buffalo.edu} \\
% Department of Computer Science,
  % University at Buffalo, State University of New York, USA
  University at Buffalo, State University of New York, USA \\
  % {\tt\small \{jsun39, rohanjag, vishnulo, changyou\}@buffalo.edu}
}

%\cortext[aaa]{Corresponding Author}

\maketitle

%%%%%%%%% ABSTRACT
\begin{abstract}
%Prompt Tuning has emerged as a prominent research paradigm for adapting vision-language models to various downstream tasks. However, existing prompt tuning methods often lead to overfitting and do not adequately regularize or align the feature spaces. To address this issue, we propose a Cross-Modal Feature Alignment method. More precisely, we introduce ways to sample certain static and stochastic anchors and optimize the prompt using the relative representations of these anchors. Our findings indicate that both types of anchors contribute to a more aligned feature space. Moreover, applying Maximum Mean Discrepancy to this aligned feature space further addresses out-of-distribution tasks. Our experiments across four different prompt network structures and 15 downstream datasets demonstrate consistent performance improvements, with gains of up to 6.1\% in Base-to-Novel generalization tasks and up to 2.2\% in out-of-distribution tasks. The code is available \href {https://anonymous.4open.science/r/promptscr-4C5B/README.md}{ here}.
Prompt Tuning has emerged as a prominent research paradigm for adapting vision-language models to various downstream tasks. However, recent research indicates that prompt tuning methods often lead to overfitting due to limited training samples. In this paper, we propose a \textbf{Cr}oss-modal \textbf{A}ligned \textbf{F}eature \textbf{T}uning (\textbf{CRAFT}) method to address this issue. Cross-modal alignment is conducted by first selecting anchors from the alternative domain and deriving relative representations of the embeddings for the selected anchors. Optimizing for a feature alignment loss over anchor-aligned text and image modalities creates a more unified text-image common space. Overfitting in prompt tuning also deteriorates model performance on out-of-distribution samples. To further improve the prompt model's robustness, we propose minimizing Maximum Mean Discrepancy (MMD) over the anchor-aligned feature spaces to mitigate domain shift. The experiment on four different prompt tuning structures consistently shows the improvement of our method, with increases of up to $6.1\%$ in the Base-to-Novel generalization task,  $5.8\%$ in the group robustness task, and $2.7\%$ in the out-of-distribution tasks. The code is available at 
%\href{https://anonymous.4open.science/r/promptscr-4C5B/README.md}{here}.
\href{https://github.com/Jingchensun/Craft}{https://github.com/Jingchensun/Craft}.%\footnotetext{Preprint. Under Review.}
\end{abstract}

%%%%%%%%% BODY TEXT
\section{Introduction}
\label{sec:intro}

Large-scale Vision Language (VL) models, such as CLIP \cite{clip}, ALIGN \cite{jia2021scaling} and LLaVA \cite{llava}, demonstrate remarkable capabilities in recognition and representation. 
%These models leverage self-supervised contrastive loss and are pre-trained on extensive image-text pairs collected from the web, covering a broad range of visual concepts and textual knowledge. Integrating these pretrained VL models into various downstream tasks \cite{od-gu2021open, od-zang2022open, od-zhou2022detecting, downstream-seg, downstream-video, krizhevsky2017imagenet, segclip1, segclip2, segclip3} holds significant promise. 
To leverage the powerful zero-shot recognition capabilities of vision-language models while avoiding the significant computational resource consumption of fine-tuning, various parameter-efficient fine-tuning methods have been proposed. Among these, prompt tuning methods \cite{prompt1, prompt2, coop, CoCoOp, vpt, visual-prompt, maple, promptsrc} have gained considerable attention in the research community due to their simple structure and relatively intuitive interpretability.

While prompt tuning methods can effectively adapt the vision-language (VL) models to various downstream tasks, a significant challenge remains unresolved: \textit{Existing prompt tuning method often lead to overfitting} \cite{overfiting-1, overfiting-2, overfiting-3, overfiting-4, overfiting-ma2023}. Existing methods optimize prompts using text-based cross-entropy loss. Although this loss function is commonly used in many image recognition tasks, however, the cross-entropy loss alone make the model more prone to overfit the training data, especially when the available training data is limited in many downstream tasks. While the prompt model is trained on only a few samples, the soft prompts concentrate on task-specific knowledge, easily over-fitting to the target task. E.g, In Table \ref{sota}, When CoOp \cite{coop} or MaPLe \cite{maple} is fine-tuned on the Base class, the recognition accuracy on the Novel class is even lower than that of the pre-trained CLIP model. 

We attribute this overfitting to the lack of regularization in the latent space. To address these challenges, we propose a cross-modal feature alignment method that leverages relative representations \cite{relative, relative2, relative3} to construct the static anchor and stochastic anchors in the dual model branch. %\blue{The static anchors are representative samples from each modality and are used as prototypes to match with the stochastic samples in every training batch.} 
While the stochastic anchors are dynamic samples that are stochastic and selected during every training batch, all the image stochastic samples and the text stochastic samples share a common semantic space. We show that combining the static anchors and the stochastic anchors can provide effective regularization for the learnable prompt and improve the generalization ability of the prompt model. 

Additionally, prompt overfitting is exacerbated by domain discrepancies in downstream tasks \cite{ multi-modal-learning, multi-modal-learning-2}. As existing methods primarily focus on acquiring task-specific knowledge from the source domains \cite{CoCoOp, maple, promptsrc}, they often neglect the domain distribution discrepancies from the source to target domains. To address this issue, we proposed an image-text joint distribution mean discrepancy minimization to enhance the cross-domain consistency of optimized prompts. Our empirical results show that the modified MMD method efficiently mitigates overfitting and improves the model's out-of-distribution recognition capability.

%distribution matching metrics like Maximum Mean Discrepancy (MMD) \cite{mmd-2, mmd-lawrence, mmd-mingshen} can be applied. However, due to scaling issues between image and text domains, applying MMD in multi-modal prompt tuning is non-trivial. We propose 
%an image-text joint distribution mean discrepancy minimization to enhance the cross-domain consistency of optimized prompts. Based on the aligned feature space, we simply to implement discrepancy minimization on the image-text joint distribution to enhance the cross-domain consistency. 

We conducted experiments on four different prompt tuning structures: Visual Prompt Tuning (VPT) \cite{visual-prompt}, Language Prompt Tuning (LPT) \cite{coop}, and Multi-modal Prompt Tuning with MaPLe \cite{maple} and PromptSRC \cite{promptsrc}. Our method consistently improved the average classification accuracy across three tasks. In the Base-to-Novel generalization experiments, applying our method to MaPLe improved accuracy by up to $6.1$ points. Additionally, applying our method to PromptSRC reduced the domain gap by $5.4$ points in the Waterbird group robustness task. Furthermore, when applied to ImageNet and its four variant datasets, accuracy improved by up to $2.7$ points. These experiments demonstrate the effectiveness of our proposed method in enhancing the generalization and robustness of prompt tuning.

Our contributions are summarized as follows:
(1) We propose a novel relative anchor-based cross-modal feature alignment method, which effectively regularizes prompt tuning and mitigates overfitting issues.
(2) We design a modified Maximum Mean Discrepancy (MMD) loss based on our aligned feature space, enhancing the model's ability to handle out-of-distribution data.
(3) We validate our approach on four different prompt tuning structures across three image classification tasks. The experimental results demonstrate the effectiveness and robustness of our method.

\section{Related Work}
\textbf{Prompt Tuning}. 
CoOp \cite{coop} pioneered the use of learnable prompts in the language branch to adapt vision-language models. Building on this concept, Visual Prompt Tuning (VPT) \cite{vpt} employs customized visual prompts tailored to image embeddings to enhance the distinguishability of image features. Unified Prompt Tuning (UPT) \cite{upt} learns a unified prompt for both image and text branches, aligning them in latent space. MaPLe \cite{maple} extends multimodal prompt tuning to deeper network structures, embedding learnable prompts into each layer of the transformer. Despite various prompt structure designs proposed for adaptation, the cross-entropy loss used in these methods will make the model prone to causing overfitting.
%they only considered the use of the text-based cross-entropy loss as the optimization objective. However, for downstream tasks, the number of training samples during the adaptation process is relatively small, thus 

\textbf{Out-of-Distribution Adaption}. Many works show that the overfitting is harmful for the out-of-distribution generalization task \cite{ dream-ood, clipood, bayesian-prompt, logoprompt, kgcoop, prograd, oegn, tpt, diverse-tpt}. To solve this problem, PromptSRC \cite{promptsrc} proposes a Gaussian weighted sampling of prompts learned at different epochs, DePT \cite{dept} decouples base-specific knowledge during tuning to preserve task-shared knowledge. 
%However, none of these methods consider the domain discrepancy between the source domain and the target domain.
Our method differs from these approaches by assuming that a feature space that is more aligned across modalities and domains can more effectively alleviate the problems of overfitting and out-of-distribution.

% Recent efforts in prompt tuning \cite{} to enhance the out-of-distribution (OOD) capabilities of the CLIP model can be categorized as two major categories. Firstly, methods aim to utilize the task-agnostic pretrain knowledge.
% Prompt Weighting \cite{prompt-weight} employs ChatGPT to generate template prompts and assigns weights automatically. Secondly, efforts focus on improving text prompt diversity. HPT \cite{hpt} constructs graphs for category-related descriptions using large language models. While existing works aim to design better prompt tuning methods based on source domain data distribution to improve the decision boundary, our approach differs in that we propose a cross-domain discrepancy match method, offering a more effective solution to the out-of-distribution problem. 

\begin{figure*}[htb!]
  \centering
  \includegraphics[width=\textwidth]{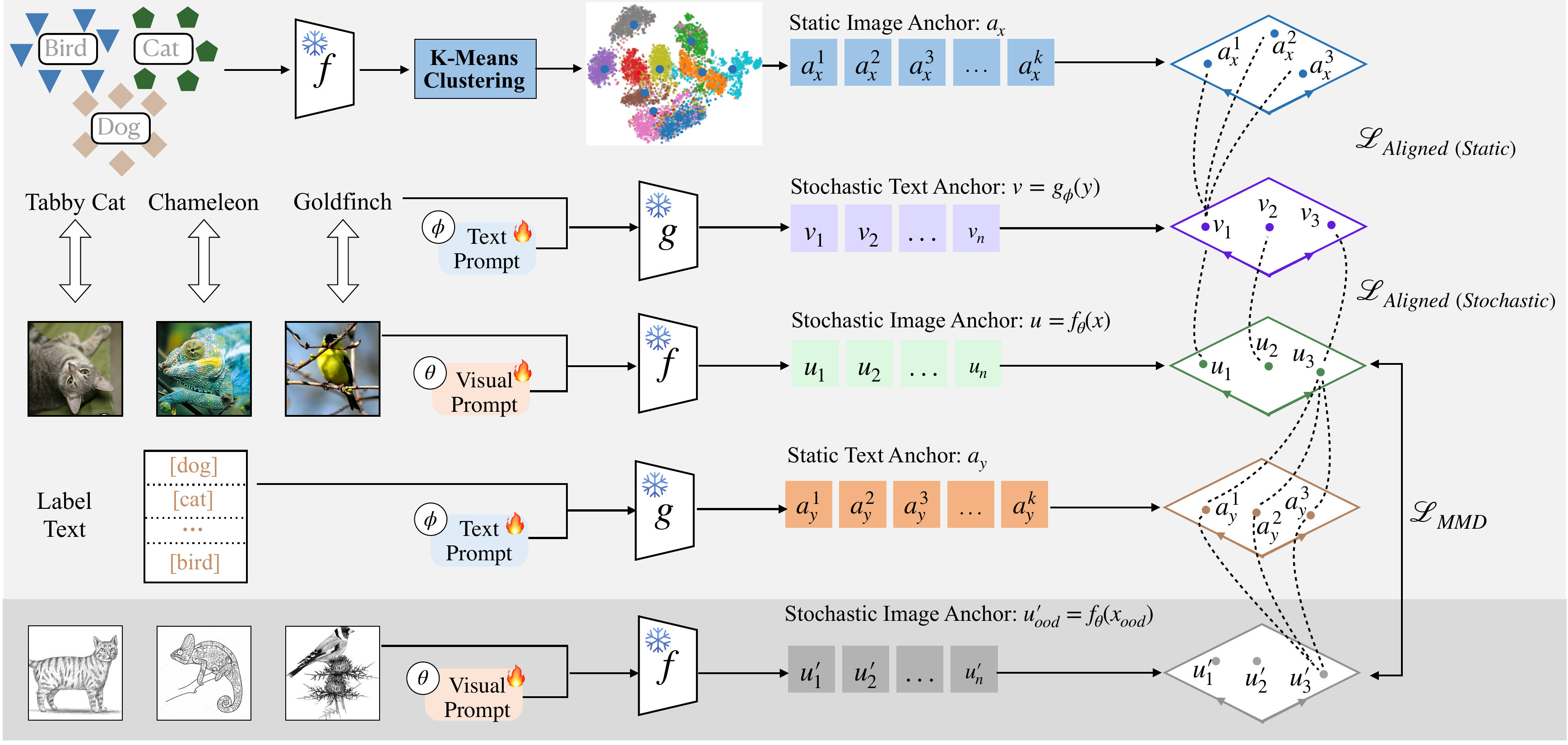}
  \caption{\textbf{Illustration of Our Proposed Cross-Modal Feature Alignment Method.} Firstly k-means clustering is conducted image embeddings to obtain static image anchors. Static text anchors are derived from the class-text labels. Simultaneously, we construct batch-level image and text samples to create stochastic image and text anchors. Static image anchors are aligned with stochastic text anchors using Equation \ref{align-image}. Additionally, stochastic text anchors and stochastic image anchors are aligned with each other by Equation \ref{l-aligned}. To address out-of-distribution samples, we apply the Maximum Mean Discrepancy (MMD) method to the aligned features, ensuring consistency within the latent space.}
  \label{fig:main}
\end{figure*}
 
\section{Method}\label{sec:method} %\subsection{Preliminary}
\textbf{Notation.} In the prompt-tuning of the vision-language models, we consider images denoted by $x$ and text by $y$. Transpose of the image vector is denoted as $x^T$. Images are drawn from a probability distribution $\P_x$ defined over measurable space $(\Omega_x,\mathcal{F}_x)$, while the text is drawn from a probability distribution $\P_y$ defined on a measurable space $(\Omega_y,\mathcal{F}_y)$. Each image-text pair is associated with a label $c$, which belongs to one of the $K$ classes. In our approach, we utilize a CLIP \cite{clip} model, which provides both an image encoder $f_\theta$ and a text encoder $g_\phi$. These encoders effectively align the feature spaces of images and text \cite{relative}.

\textbf{Incorporating Prompt Tuning Parameters into $\{\theta, \phi\}$.} In prompt-tuning methods, certain learnable prompts are prepended to the text prompts onto the text branch, denoted by $\Lambda_\text{text} \in \left\{{\lambda}_{\text{text}}^1, {\lambda}_{\text{text}}^2, \cdots, {\lambda}_{\text{text}}^L\right\} $. Similarly, learnable prompts $\Lambda_\text{visual} \in \left\{{\lambda}_{\text{visual}}^1, {\lambda}_{\text{visual}}^2, \cdots, {\lambda}_{\text{visual}}^L\right\} $,  are prepended into the image branch. For more details regarding the process of prepend, refer to MaPLe \cite{maple} architecture or Figure~\ref{fig:main} in the paper. In an image classification task, the tokenized image embeddings and learnable visual prompts are fed into the image encoder \cite{maple}, resulting in image features $f_{{\theta}}\left({x, {\Lambda}_{\text{visual}}} \right)$. Similarly, for the text encoder, tokenized class embeddings and learnable text prompts are processed in the text encoder, yielding text embeddings $g_{{\phi}}\left({y}, {{\Lambda}_{\text{text}}} \right)$. To maintain simplicity in this paper, the notation $\{\theta, \phi\}$ will be used to represent the parameters that include $\{{\Lambda}_{\text{visual}}, {\Lambda}_{\text{text}}\}$ respectively.

\textbf{Our Feature Alignment Method.} Existing methods mainly use text-based cross-entropy loss, which leads to overfitting, as elaborated in Section~\ref{sec:intro}. We propose an anchor-based Cross-modal Feature Alignment method. Our methodology involves generating a distribution over the query point $x$ (or $y$) by applying a softmax function to the distances from certain anchors in the alternate modality. Cross-modal alignment is possible by comparing samples from the alternative domain (say image) for every sample in the given domain (say text). A distribution of anchors $\Ax_X$ from $\P_X$ and another discrete distribution $\Ax_Y$  from $\P_Y$ can be drawn to define the anchors. Comparisons made on these anchors help in reducing computational complexity; instead of comparing them to all, we only compare them to a few. Having defined anchors, let us write down the distribution of classes for every query image-text pair $(x, y)$. For $a_x \sim \Ax_X$ and $a_y \sim \Ax_Y$ and using $\langle \cdot, \cdot \rangle$ as the inner product measure, we have the discrete probability density function over the image/text domains defined as

\begin{align}
p_{x}(c = k \mid x; \theta) &= \frac{\exp \langle f_{\theta}(x), a_{y}^{k} \rangle}{\sum_{k'} \exp \langle f_{\theta}(x), a_{y}^{k'} \rangle}  \label{align-text} \\
p_{y}(c = k \mid y; \phi) &= \frac{\exp \langle g_{\phi}(y), a_{x}^{k} \rangle}{\sum_{k'} \exp \langle g_{\phi}(y), a_{x}^{k'} \rangle}  \label{align-image}
\end{align}

Minimizing the negative log-probability of Equation \ref{align-text} and Equation \ref{align-image}, we obtain the feature alignment loss as follows:

\begin{equation}
 \mathcal{L}_{\text{Aligned}}(\theta, \phi)=-\log{p_x(c=k \mid x; \theta)p_y(c=k \mid y; \phi)} \\
 \label{l-aligned}
\end{equation}
Minimizing negative log probability often involves transitioning to expectations, while the objective is ideally expressed as an expectation over the true data distribution, in practice, we approximate this using the empirical distribution of observed data. 
Training proceeds by minimizing $\mathcal{L}_{\text{Aligned}}$ of the true class $k$ through SGD \cite{ruder2016overview}. Such a cross-modal loss function promotes the alignment of feature spaces across both the text and image modalities \cite{norelli2024asif}. Additionally, leveraging anchors from the alternative modality can offer more reliable classification cues \cite{snell2017prototypical}.

\subsection{Anchors selection for feature alignment}
In the previous section, we have already two types of anchors - the imaging anchors, denoted as $a_x$, and the text anchors, denoted as $a_y$. Based on how we sample a given image or text anchor, we introduce two additional categories.
\begin{enumerate}
  \setlength{\itemsep}{0pt}\setlength{\parskip}{0pt}
  \setlength{\parsep}{0pt}
  \setlength{\leftmargin}{0pt}
  \item \textbf{Static Anchors.} Anchors are chosen before the training begins. The selected anchors remain the same throughout the training process. For the text modality, these anchors might represent a static template, such as ``a photo of $\{\}$" where the class-name is inserted within $\{\}$. For the image modality, static anchors can be determined using clustering algorithms.
  \item \textbf{Stochastic Anchors.} As the name implies, these anchors are chosen randomly in every iteration of training. In a given iteration, a fixed count of anchors are sampled from both the text and image modality.
\end{enumerate}
Using static anchors in \eqref{l-aligned} yields $\mathcal{L}_{\text{Aligned}}(\text{static})$. We obtain $\mathcal{L}_{\text{Aligned}}(\text{stocastic})$ using stochastic anchors in \eqref{l-aligned}. Overall, 
\begin{align}
    \mathcal{L}_{\text{Aligned}} = \mathcal{L}_{\text{Aligned}}(\text{static}) + \mathcal{L}_{\text{Aligned}}(\text{stocastic})
\end{align}
The interplay between static anchors and stochastic anchors is visible in Figure~\ref{fig:main}. %On every gradient descent step, the loss from \eqref{l-aligned} integrates contributions from both the static and stochastic anchors.\\

{\bf Necessity of Static and Stochastic Anchors. } Static anchors provide reliable reference points throughout training, ensuring consistency during feature comparisons. \blue{Stochastic anchors serve as a form of data augmentation. In each iteration, samples from various modalities are presented to the model, allowing for sampling from the complete support of the data distribution. This approach enhances robustness against minor perturbations and outliers, resulting in more stable and generalized performance. \cite{baldi2013understanding}.}

%\blue{Conversely, stochastic anchors facilitate exploration within the feature space of the domain. }Introducing variability through changing anchors, via stochastic anchors, allows models to escape local minima \cite{yu2021simple}.

\begin{remark}
Pre-trained vision-language (V-L) dual model architectures, like MaPLe \cite{maple}, employ text-based cross-entropy loss. In text cross-entropy loss, a random batch of images are drawn and are compared to static text-labels. This is simply optimizing for $\mathcal{L}_{\text{Aligned}}(\theta)\propto \langle f_{\theta}(x), a_{y} \rangle$, where $a_y$ represent the static text anchors defined by text-labels. Thus, our proposed $\mathcal{L}_{\text{Aligned}}(\theta, \phi)$ from \eqref{l-aligned} subsumes text-cross entropy loss. 
\end{remark}

\subsection{Accounting for the variations in distribution }%within samples}
Although prompt-tuning strategies enhance the performance of vision language models, they have a tendency to overfit the samples, as described in Section~\ref{sec:intro}. Overfitting leads to the generation of inaccurate and uncertain predictions by models when dealing with samples that either have different distributions or belong to Novel classes \cite{blei2017variational}. {\blue{This paper will analyze two main distributions of image data, the in-domain samples $x_\text{id} \sim \P^\text{id}_x$ and the out-of-domain samples $x_\text{ood} \sim \P^\text{ood}_x$. }
%Unless explicitly specified, $x$ is assumed to be an in-domain sample $x_\text{id}$ throughout the paper. While data can be derived from multiple distributions, we will concentrate on analyzing two distributions at a time to ensure simplicity and clarity. 
Extending multiple distributions using the concept of group robustness is achievable \cite{sohoni2021barack}.

Applying a parametric function to any of the two distributions, $x_\text{id}$ and $x_\text{ood}$, and updating the function's parameters to make $\P^\text{id}_x$ and $\P^\text{ood}_x$ indistinguishable from each other has been demonstrated to be effective in the literature \cite{gretton2012kernel}. 
%The technique is straightforward: estimate a particular metric of disparity between the distributions and tune the parameters of the function transformation in order to minimize this disparity. 
Although techniques such as KL divergence, Optimal Transport measures, can be effective in measuring distributional discrepancies, the maximum mean discrepancy (MMD) is widely favored due to its applicability to various types of data, including imaging, and its well-established statistical properties \cite{gretton2012kernel} (page $728$) and \cite{dudley2018real} (page $292$). 
%The rationale to employ MMD arises from the following result as mentioned in\cite{gretton2012kernel} (page $728$) and \cite{dudley2018real} (page $292$). 
\begin{lemma}
    \label{dudley-result}
    $(\Omega_x, \mathcal{F}_x)$ is a measurable space, and $\P^{\text{id}}_x$ and $\P^{\text{ood}}_x$ are two borel probability measures for in-domain and out-of-domain imaging data. Then  $\P^{\text{id}}_x = \P^{\text{ood}}_x$ if and only if $\E_{x_\text{id}}(f(x_\text{id})) = \E_{x_\text{ood}}(f(x_\text{ood}))$ for all $c\in C(\Omega_x)$ where $C(\Omega_x)$ is the space of bounded continuous functions on $\Omega_x$.
\end{lemma}
%In order to deal with two separate measure spaces $(\Omega_x, \mathcal{F}_x)$ and $(\Omega_y, \mathcal{F}_y)$, it may be necessary to make extra assumptions such as the use of borel isomorphism \cite{dudley2018real} (see page $487$). 
A function class that is sufficiently rich is required to uniquely determine whether $\P^{\text{id}}_x$ = $\P^{\text{ood}}_x$ as per Lemma~\ref{dudley-result}. The unit balls in Reproducing Kernel Hilbert space (RKHS) are such a function class that are utilized to establish maximum mean discrepancy (MMD) \cite{gretton2012kernel}.

\blue{MMD measures the distributional discrepancy between in-domain and out-of-domain data; however, challenges occur with multi-modal data, involving vision and language. 
%Evaluating MMD over the joint distribution of the data, represented as $\P_x \times \P_y$, presents significant challenges due to scaling issues. 
The straightforward use of MMD necessitates optimization across vision-language, language-language, and vision-vision distribution pairs. We propose calculating the Maximum Mean Discrepancy (MMD) through a vision-language similarity measure. }
%The problem then lies in incorporating the text probability measure $\P_y$ to minimize MMD between $\P^{\text{id}}_x$ and $\P^{\text{ood}}_x$. }
An obvious solution is to utilize a product measure $\P_x \times \P_y$. The product measure theorem (Theorem~$4.4.6$ from \cite{dudley2018real}) enables the use of product measures because there exists a unique measure on the product of the $\sigma-$algebras $\mathcal{F}_x \times \mathcal{F}_y$. It reminds us of the fundamental \cite{dudley2018real}, 
\begin{fact}
    \label{product-fact}
The cardinality of the $\sigma-$algebra for the product measure is greater than that of the individual probability measures.
\end{fact}
Please refer to page $139$ of \cite{dudley2018real} for more details. \blue{The above Fact~\ref{product-fact} significantly affects MMD computation, as empirical estimates require more samples now to approximate expectation calculations. As a solution, we propose utilizing an induced measure to evaluate MMD in lieu of evaluating it over product measures.} For the image encoder $f_\theta$ and a unit-normed text anchor $a_y$, let $a_y^T f_\theta(\cdot)$ be a measurable function from $\Omega_x, \mathcal{F}_x$ to $\Gamma_x, \mathcal{G}_x$. Our induced measure, the anchor-aligned probability measure, is  

\begin{align}
    \P_x^{a_y} := \P_x \circ (a_y^T f_\theta)^{-1}(B) = \P_x\big((a_y^T f_\theta)^{-1}(B)\big), B\in\mathcal{G}
    \label{eq5}
\end{align}
%The anchor-aligned probability measure $\P_x^{a_y}$ produces anchor aligned-feature vectors and anchor-aligned MMD measure defined as follows,

\blue{ Based on the the anchor-aligned probability measure $\P_x^{a_y}$ defined in Equation \ref{eq5}, the anchor-aligned MMD measure defined as follows:}

%An important difficulty we face when dealing with multi-modal data, namely combining vision and language, is that evaluating MMD (Maximum Mean Discrepancy) over the joint distribution of the data, denoted as $\P_x \times \P_y$, is not straightforward due to scaling issues. To address this issue, we will calculate the Maximum Mean Discrepancy (MMD) using the anchor-aligned feature vectors. Therefore, we define our anchor-aligned MMD score as follows:

\begin{definition}
   Given a non-negative, characteristic, and bounded kernel $k$ in a Reproducing Kernel Hilbert Space (RKHS), with a bounded search region $\theta$, and a holder-continuous function $f_\theta$ with a specific ratio and support, we consider a text anchor $a_y$ that belongs to the same class $c=k$ as the image embedding vector $x$. We calculate the maximum mean discrepancy between anchor-aligned feature vectors from two distributions, $\P_x^\text{id}$ and $\P_x^\text{ood}$ as
    \begin{align}
        \mathcal{L}_{\text{MMD}} = \| \E_{x\sim\P_x^\text{id}} k(f_\theta^T(x)a_y, \cdot) - \E_{x\sim\P_x^\text{ood}} k(f_\theta^T(x)a_y, \cdot) \|
        \label{eq6}
    \end{align}
\end{definition}
\blue{Equation \ref{eq6} defines the MMD loss between two distributions $\P^{\text{id}}_x$ (in-domain) and $\P^{\text{ood}}_x$ (out-of-domain) over the anchor-aligned feature space.}
\blue{Here, $k(x,\cdot)$ signifies that the kernel has one parameter fixed at $x$, while the second parameter, represented by $\cdot$, is free and can accommodate any random variable.} In particular, with the feature mapping onto RKHS as $\Phi$, we have $\langle \Phi(x), \Phi(y) \rangle = k(x,y)$. In the empirical form of the MMD, the expectation is replaced with the sample average calculated over a group. A previous study \cite{zhou2018statistical} is of particular relevance because it has demonstrated that calculating the average of samples taken from training batches ( sub-samples) is feasible while maintaining consistency and asymptotic normality criteria. Moreover, \cite{zhou2018statistical} can be seen as an extension of the paper \cite{baktashmotlagh2013unsupervised} where the in-domain and out-domain samples are parameterized separately.
For a detailed explanation of the consistency and asymptotic properties of MMD and the necessary assumptions, we recommend the reader to visit \cite{zhou2016hypothesis}. However, it is crucial that the anchor-aligned feature vectors require holder continuity. We demonstrate the validity of the statement below

\begin{lemma}
    \label{lemma-holder}
If $f_\theta$ exhibits Holder continuity with a constant $\alpha$ and the anchor vectors are normalized to have unit length, then the anchor-aligned feature vectors likewise exhibit Holder continuity with the same constant $\alpha$.\\
    {\bf Proof Outline. } 
\begin{align}
    \|f_\theta(x_\text{id})^T a_y - f_\theta(x_\text{ood})^T a_y\| 
    &\le \|f_\theta(x_\text{id}) - f_\theta(x_\text{ood})\| \|a_y\| \nonumber \\
    &\le \blue{\|x_\text{id} - x_\text{ood}\|^\alpha}
    \label{eq7}
\end{align}

\end{lemma}
 \blue{The first inequality in Equation \eqref{eq7} is valid by the Cauchy-Schwarz inequality. The second inequality is valid because of the unit norm constraint on the anchor aligned vectors $a_y$.} The lemma~\ref{lemma-holder} ensures that the consistency properties and the asymptotic guarantees will remain valid for the feature vectors that are aligned with the anchor.

\section{Experiment}
In order to assess the effectiveness of our proposed method, we conducted three primary experiments: 1) Base-to-Novel Generalization on 11 datasets, 2) Cross-Domain Robustness on Waterbird and CelebA dataset, 3) Out-of-Distribution Task on ImageNet and its four variant datasets.

\subsection{Feature Alignment Improves the Base-to-Novel Generalization}

%\definecolor{LightCyan}{rgb}{0.88,1,1}
% \definecolor{LightRed}{rgb}{1,0.88,0.88}
% \definecolor{Gray}{rgb}{0.8,0.8,0.8}
% \definecolor{LightGray}{rgb}{0.92,0.92,0.92}
% \definecolor{LightPurple}{RGB}{226, 225, 254}

\newcolumntype{G}{>{\columncolor{LightCyan}}c}
% \usepackage{multirow}
\begin{table*}[htp!]
\centering
\setlength{\tabcolsep}{1.2pt}
\fontsize{7}{8}\selectfont
\newcommand{\fivept}{\fontsize{5}{4}\selectfont\color{black}}
\newcommand{\sixept}{\fontsize{6}{5}\selectfont}
\newcommand{\redpt}{\fontsize{7}{8}\selectfont \color{blue}}
\begin{tabular}{@{}c|cGcGcGcGcGcGcGcGcGcGcG|cG@{}}
\toprule
Method                  & \multicolumn{2}{c}{OXfordPets}      & \multicolumn{2}{c}{Flowers102}   & \multicolumn{2}{c}{FGVC-Aircraft}  & \multicolumn{2}{c}{DTD}       & \multicolumn{2}{c}{EuroSAT}   & \multicolumn{2}{c}{StanfordCars}      & \multicolumn{2}{c}{Food101}      & \multicolumn{2}{c}{Caltech101}   & \multicolumn{2}{c}{UCF101}       & \multicolumn{2}{c}{SUN397}       & \multicolumn{2}{c}{ImageNet}  & \multicolumn{2}{c}{Average}       \\ \midrule
                        & Base          & Novel         & Base          & Novel         & Base          & Novel         & Base          & Novel         & Base          & Novel         & Base          & Novel         & Base          & Novel         & Base          & Novel         & Base          & Novel         & Base          & Novel         & Base          & Novel         & Base          & Novel         \\
\cmidrule(lr){2-3} \cmidrule(lr){4-5} \cmidrule(lr){6-7} \cmidrule(lr){8-9} \cmidrule(lr){10-11} \cmidrule(lr){12-13} \cmidrule(lr){14-15} \cmidrule(lr){16-17} \cmidrule(lr){18-19} \cmidrule(lr){20-21} \cmidrule(lr){22-23}  \cmidrule(lr){24-25} 
LPT \cite{maple}                    & 94.8          & 97.3          & 96.4          & 73.6          & 33.0          & 28.3          & 79.4          & 53.1          & 87.1          & 68.9          & 72.6          & 74.0          & 90.2          & 91.2          & 97.9          & 93.7          & 84.2          & 75.7          & 79.3          & 77.3          & 76.2          & 70.8          & 81.0          & 73.1          \\
\textbf{Ours-LPT}         & 96.1       & 97.9        & 97.6         & 75.2         & 37.1         & 33.0          & 81.1       & 56.4       & 89.3         & 70.9         & 73.6       & 75.1        & 90.5        & 91.7       & 98.1         & 95.9         & 84.5       & 78.7       & 79.9       & 78.1       & 76.6         & 71.1          & 82.2       & 74.9       \\
\rowcolor{LightGray}
$\bigtriangleup$ & +1.3       & +0.6        & +1.2         & +1.6         &\cellcolor{LightPurple} +4.1         & \cellcolor{LightRed}+4.7          & +1.7       & +3.3       & +2.2         & +2.0         & +1.0       & +1.1        & +0.3        & +0.5       & +0.2         & +2.2         & +0.3       & +3.0       & +0.6       & +0.8       & +0.4         & +0.3          & \textbf{+1.2}       & \textbf{+1.8 }      \\ \midrule
VPT \cite{visual-prompt}                    & 94.8          & 96.1          & 85.2          & 68.5          & 30.8          & 33.8          & 77.7          & 53.3          & 89.4          & 69.0          & 68.8          & 73.4          & 89.3          & 90.1          & 97.0          & 93.9          & 81.9          & 73.1          & 75.7          & 77.7          & 74.8          & 69.2          & 78.7          & 72.6          \\
\textbf{Ours-VPT}         & 95.3       & 96.5        & 90.0         & 73.2         & 34.6         & 34.7          & 80.2       & 57.1       & 92.9         & 74.5         & 70.0       & 74.8        & 90.4        & 91.6       & 98.1         & 94.2         & 82.9       & 75.3       & 76.2       & 77.8       & 75.3         & 69.4          & 80.5       & 74.5       \\
\rowcolor{LightGray}
$\bigtriangleup$ & +0.5       & +0.4        &\cellcolor{LightPurple} +4.8         & +4.7         & +3.8         & +0.9          & +2.5       & +3.8       & +3.5         &\cellcolor{LightRed} +5.5         & +1.2       & +1.4        & +1.1        & +1.5       & +1.1         & +0.3         & +1.0       & +2.2       & +0.5       & +0.1       & +0.5         & +0.2          & \textbf{+1.9}       & \textbf{+1.9}       \\ \midrule
MaPLe \cite{maple}                  & 95.4          & 96.4          & 95.9          & 72.5          & 37.4          & 35.6          & 80.4          & 59.2          & 94.1          & 75.2          & 72.9          & 72.2          & 90.7          & 90.7          & 97.7          & 93.6          & 83.0          & 75.4          & 80.8          & 76.0          & 76.7          & 69.1          & 82.3          & 74.2          \\
\textbf{Ours-MaPLe }      & 95.5       & 98.0        & 97.8         & 75.3         & 42.3         & 37.1          & 81.7       & 63.9       & 96.3         & 78.1         & 79.0       & 74.5        & 89.9        & 92.2       & 98.7         & 95.2         & 86.7       & 78.6       & 81.5       & 78.7       & 77.3         & 70.1          & 84.2       & 76.5       \\
\rowcolor{LightGray}
$\bigtriangleup$ & +0.1       & +1.6        & +1.9         & +2.8         & +4.9         & +1.5          & +1.3       &\cellcolor{LightRed} +4.7       & +2.2         & +2.9         &\cellcolor{LightPurple} +6.1       & +2.3        & -0.8       & +1.4       & +1.0         & +1.6         & +3.7       & +3.2       & +0.7       & +2.7       & +0.6         & +1.1          & \textbf{+2.0}       & \textbf{+2.3 }      \\ \midrule
PromptSRC \cite{promptsrc}              & 95.0          & 97.4          & 98.0          & 78.0          & 42.1          & 37.0          & 82.4          & 59.3          & 91.4          & 73.9          & 77.9          & 75.2          & 90.6          & 91.7          & 97.9          & 94.2          & 86.2          & 78.1          & 81.7          & 78.4          & 77.2          & 70.5          & 83.7          & 75.8          \\
\textbf{Ours-PromptSRC}   & 95.8       & 98.0        & 98.7         & 77.7         & 47.0         & 38.2          & 84.4       & 64.0       & 95.5         & 79.4         & 80.8       & 75.6        & 91.0        & 92.2       & 98.1         & 95.3         & 88.6       & 80.5       & 83.2       & 78.9       & 78.2         & 70.8          & 85.6       & 77.3       \\
\rowcolor{LightGray}
 $\bigtriangleup$ & +0.8       & +0.6        & +0.7         & -0.3        &\cellcolor{LightPurple} +4.9         & +1.2          & +2.0       & +4.7       & +4.1         &\cellcolor{LightRed} +5.5         & +2.9       & +0.4        & +0.4        & +0.5       & +0.2         & +1.1         & +2.4       & +2.4       & +1.5       & +0.5       & +1.0         & +0.3          & \textbf{+1.9}       & \textbf{+1.5}                \\ \bottomrule
\end{tabular}
\caption{\textbf{Base-to-Novel Generalization on 11 Datasets.} In this setting, the prompt model is trained on Base group data and evaluated on Novel group data. Our method differs from the baseline in that we apply $\mathcal{L}_{\text{Aligned}}$ to the baseline method. \colorbox{LightPurple}{\makebox[0.04in][c]{\rule{0pt}{0.04in}}} represents the max gains in the Base group, \colorbox{LightRed}{\makebox[0.04in][c]{\rule{0pt}{0.04in}}} represents the max gains in the Novel group. The table shows that our feature alignment method consistently improves the average classification accuracy across four different prompt-tuning structures.}
\label{table1}
\end{table*}

\begin{table*}[htp!]
\centering
\setlength{\tabcolsep}{1.5pt}
\fontsize{7}{8}\selectfont
\newcommand{\fivept}{\fontsize{5}{4}\selectfont\color{black}}
\newcommand{\sixept}{\fontsize{6}{5}\selectfont}
\newcommand{\redpt}{\fontsize{7}{8}\selectfont \color{blue}}
\begin{tabular}{@{}c|cGcGcGcGcGcGcGcGcGcGcG|cG@{}}
\toprule
Method                  & \multicolumn{2}{c}{OXfordPets}      & \multicolumn{2}{c}{Flowers102}   & \multicolumn{2}{c}{FGVC-Aircraft}  & \multicolumn{2}{c}{DTD}       & \multicolumn{2}{c}{EuroSAT}   & \multicolumn{2}{c}{StanfordCars}      & \multicolumn{2}{c}{Food101}      & \multicolumn{2}{c}{Caltech101}   & \multicolumn{2}{c}{UCF101}       & \multicolumn{2}{c}{SUN397}       & \multicolumn{2}{c}{ImageNet}  & \multicolumn{2}{c}{Average}       \\ \midrule
                        & Base          & Novel         & Base          & Novel         & Base          & Novel         & Base          & Novel         & Base          & Novel         & Base          & Novel         & Base          & Novel         & Base          & Novel         & Base          & Novel         & Base          & Novel         & Base          & Novel         & Base          & Novel         \\
\cmidrule(lr){2-3} \cmidrule(lr){4-5} \cmidrule(lr){6-7} \cmidrule(lr){8-9} \cmidrule(lr){10-11} \cmidrule(lr){12-13} \cmidrule(lr){14-15} \cmidrule(lr){16-17} \cmidrule(lr){18-19} \cmidrule(lr){20-21} \cmidrule(lr){22-23}  \cmidrule(lr){24-25}
CLIP \cite{clip}    & 91.2 & 97.3 & 72.1 & 77.8 & 27.2 & 36.3 & 53.2 & 59.9 & 56.5 & 64.1 & 63.4 & 74.9 & 90.1 & 91.2 & 96.8 & 94.0 & 70.5 & 77.5 & 69.4 & 75.4 & 72.4 & 68.1 & 69.3 & 74.2 \\
CoOp \cite{coop}            & 93.7          & 95.3          & 97.6          & 59.7          & 40.4          & 22.3          & 79.4          & 41.2          & 92.2          & 54.7          & 78.1          & 60.4          & 88.3          & 82.3          & 98.0          & 89.8          & 84.7          & 56.1          & 80.6          & 65.9          & 76.5          & 67.9          & 82.7          & 63.2          \\
Co-CoOp \cite{CoCoOp}         & 95.2          & 97.7          & 94.9          & 71.8          & 33.4          & 23.7          & 77.0          & 56.0          & 87.5          & 60.0          & 70.5          & 73.6          & 90.7          & 91.3          & 98.0          & 93.8          & 82.3          & 73.5          & 79.7          & 76.9          & 76.0          & 70.4          & 80.5          & 71.7          \\
KgCoOp \cite{kgcoop}          & 94.7          & 97.8          & 95.0          & 74.7          & 36.2          & 33.6          & 77.6          & 55.0          & 85.6          & 64.3          & 71.8          & 75.0          & 90.5          & 91.7          & 97.7          & 94.4          & 82.9          & 76.7          & 80.3          & 76.5          & 75.8          & 70.0          & 80.7          & 73.6          \\
ProGrad \cite{prograd}         & 95.1          & 97.6          & 95.5          & 71.9          & 40.5          & 27.6          & 77.4          & 52.4          & 90.1          & 60.9          & 77.7          & 68.6          & 90.4          & 89.6          & 98.0          & 93.9          & 84.3          & 74.9          & 81.3          & 74.2          & 77.0          & 66.7          & 82.5          & 70.7          \\
HPT \cite{hpt}             & 95.8          & 97.7          & 98.2          & 78.4          & 42.7          & 38.1          & 83.8          & 63.3          & 94.2          & 77.1          & 77.0          & 74.2          & 90.5          & 91.6          & 98.4          & 95.0          & 86.5          & 80.1          & 82.6          & 79.3          & 78.0          & 70.7          & 84.3          & 76.9          \\
OGEN \cite{ogen}            & 96.0          & 97.5          & 97.3          & 77.7          & 41.3          & 40.3          & 83.8          & 62.5          & 93.4          & 76.7          & 77.6          & 75.2          & 90.7          & 91.7          & 98.3          & 94.8          & 87.4          & 79.3          & 82.6          & 78.8          & 77.5          & 71.0          & 84.2          & 76.9          \\
DePT \cite{dept}            & 95.4          & 97.3          & 98.4          & 77.1          & 45.7          & 36.7          & \textbf{84.8}          & 61.2          & 93.2          & 77.9          & 80.8          & 75.0          & 90.9          & 91.6          & \textbf{98.6}          & 94.1          & 87.7          & 77.7          & 83.3          & \textbf{79.0}          & 78.2          & 70.3          & 85.2          & 76.2          \\ \midrule
\textbf{Ours-Best Model}    & \textbf{95.8} & \textbf{98.0} & \textbf{98.7} & \textbf{77.7} & \textbf{47.0} & \textbf{38.2} & 84.4 & \textbf{64.0} & \textbf{95.5} & \textbf{79.4} & \textbf{80.8} & \textbf{75.6} & \textbf{91.0} & \textbf{92.2} & 98.1 & \textbf{95.3} & \textbf{88.6} & \textbf{80.5} & \textbf{83.2} & 78.9 & \textbf{78.2} & \textbf{70.8} & \textbf{85.6} & \textbf{77.3} \\ 
\rowcolor{LightGray}
$\bigtriangleup$ & +0.4          & +0.7          & +0.3          & +0.6          & +1.3          & +1.4          & -0.4         &\cellcolor{LightRed} +2.8          &\cellcolor{LightPurple}  +2.3          & +1.5          & +0.0          & +0.6          & +0.1          & +0.6          & -0.5         & +1.2          & +0.8          &\cellcolor{LightRed} +2.8          & +0.0          & -0.1         & +0.0          & +0.6          & \textbf{+0.4 }         &\textbf{+1.2}          \\ \bottomrule
\end{tabular}
\caption{\textbf{Ours Best Model Compare with Some State-of-art Methods.} Our best model here is the PromptSRC model comes from Table \ref{table1}. When compared to the state-of-the-art method, our feature alignment method mitigates the overfitting issues on novel classes, thus leading to higher accuracy even than the SOTA method.}
\label{sota}
\end{table*}

{\bf Setting.} We utilize 11 publicly available image classification datasets for experiments, including ImageNet \cite{imagenet}, Caltech101 \cite{caltech101}, OxfordPet \cite{oxfordpets}, Flowers102 \cite{flowers102}, StanfordCars \cite{standfordcars}, Food101 \cite{food101}, FGVC-Aircraft \cite{fgv}, SUN397 \cite{sun397}, DTD \cite{dtd}, EuroSAT \cite{eurosat}, and UCF101 \cite{ucf101}. Each dataset is split into non-overlapping Base and Novel classes. We fine-tune the model in Base classes and evaluate it in Novel classes, we call this setting Base-to-Novel Generalization. The results are displayed in Table \ref{table1} and Table \ref{sota}.
%A key take-away of Table \ref{table1} The introduction of $\mathcal{L}_{\mathrm{Aligned}}$ consistently enhances the Base group performance across four prompt tuning methods. 

\textbf{The feature alignment method improves the accuracy of both Base and Novel group classes.} In Table \ref{table1}, for the Base group, the final average results were consistently enhanced by applying $\mathcal{L}_{\text{Aligned}}$ to four prompt tuning structures, with a $2.0$-point gain achieved by MaPLe \cite{maple} and a $1.9$-point gain achieved by PromptSRC \cite{promptsrc} on average. For some individual datasets, applying our method $\mathcal{L}_{\mathrm{Aligned}}$ to MaPLe improved the results by 6.1 points on StanfordCars. 
%while applying it to PromptSRC led to a 4.9-point improvement on FGVC-Aircraft.
Additionally, our $\mathcal{L}_{\text{Aligned}}$ loss consistently enhanced performance on novel classes. An average gain of $2.3$ points is yielded by $\mathcal{L}_{\text{Aligned}}$ on MaPLe, while an average improvement of $1.5$ points is resulted on PromptSRC. 
%The most significant gain on individual datasets was $5.5$ points on EuroSAT for PromptSRC. 
Although some datasets, such as Flowers102 and Food101, show only minimal improvements, we attribute this to the already high recognition accuracy (over $90\%$) in these tasks from the original CLIP model, thus the improved space is not as significant as other tasks (like DTD or EuroSAT).  In most datasets, better results are achieved by our method compared to the baseline models.

\textbf{The best model within feature aligned is also very competitive among state-of-the-art methods.} From table \ref{sota}, it can be seen that while existing prompt tuning methods (E.g., CoOp \cite{coop}, Co-CoOp \cite{CoCoOp}, ProGrad \cite{prograd}, etc.) perform well on the base class, they perform poorly on the novel class, with recognition accuracy even lower than the original CLIP model. In contrast, our best model achieves better accuracy than the original CLIP model, which indicates the proposed feature alignment method can prevent overfitting on the prompt model. When compared our best approach, namely $\mathcal{L}_{\mathrm{Aligned}}$ applied on PromptSRC from Table \ref{table1}, with state-of-the-art methods like KgCoOp \cite{kgcoop}, ProGrad \cite{prograd}, HPT \cite{hpt}, OGEN \cite{oegn}, and DePT \cite{dept}. It is seen that our best model maintains the same recognition level on the base group (with a $0.4$-point improvement), and outperforms the state-of-the-art prompt-tuning methods by $1.4$ points on the novel class. The better performance on most individual datasets is also achieved by our method, with a maximum gain of $2.3$ points on the Base group and $2.8$ points on the Novel group. These results further demonstrate that our method is highly competitive with state-of-the-art methods.

\begin{table}[htb!]
\begin{minipage}[t]{0.45\textwidth}
\setlength{\tabcolsep}{2pt}
\centering
\fontsize{7}{7}\selectfont
\newcommand{\fivept}{\fontsize{4}{4}\selectfont\color{black}}
\newcommand{\sixept}{\fontsize{6}{5}\selectfont}
\newcommand{\redpt}{\fontsize{7}{8}\selectfont \color{blue}}
\begin{tabular}{@{}c|cccccccc@{}}
\toprule
                                         & \multicolumn{4}{c}{Waterbirds}                              & \multicolumn{4}{c}{CelebA}                                 \\ 
  \cmidrule(lr){2-5} \cmidrule(lr){6-9}                                        
                                          & WG             & Avg           & Gap          &$\bigtriangleup$         & WG            & Avg           & Gap          & $\bigtriangleup$          \\
                                               \midrule
ERM Linear Probe  \cite{kumar2022fine}                        & 7.9            & 93.5          & 85.6         &               & 11.9          & 94.7          & 82.8         &               \\
ERM Adapter   \cite{clip-adapter}                            & 60.8           & 96.0          & 35.2         &               & 36.1          & 94.2          & 58.1         &               \\
WiSE-FT \cite{wortsman2022robust}                                  & 49.8           & 91.0          & 41.2         &               & 85.6          & 88.6          & 3.0          &               \\
DFR (Subsample) \cite{kirichenko2022last}                          & 63.9           & 91.8          & 27.9         &               & 76.9          & 92.5          & 15.6         &               \\
DFR (Upsample)  \cite{zhang2022contrastive}                          & 51.3           & 92.4          & 41.1         &               & 89.6          & 91.8          & 2.2          &               \\
Contrastive Adapter \cite{zhang2022contrastive}                      & 83.7           & 89.4          & 5.7          &               & 90.0          & 90.7          & 0.7          &               \\
\toprule
LPT \cite{maple}                                      & 78.5           & 89.6          & 11.1         &               & 78.8          & 88.6          & 9.8          &               \\
\textbf{Ours-LPT}       & \textbf{79.2}  & \textbf{88.8} & \textbf{9.6} & \cellcolor{LightRed} \textbf{-1.5} $\downarrow$   & \textbf{80.4} & \textbf{87.9} & \textbf{7.5} & \cellcolor{LightRed} \textbf{-2.3} $\downarrow$ \\
\midrule
VPT \cite{visual-prompt}                                      & 79.3           & 87.7          & 8.4          &               & 79.9          & 89.8          & 9.9          &               \\
\textbf{Ours-VPT}         & \textbf{82.1}  & \textbf{88.9} & \textbf{6.8} & \cellcolor{LightRed} \textbf{-1.6} $\downarrow$  & \textbf{82.2} & \textbf{90.4} & \textbf{8.2} & \cellcolor{LightRed}  \textbf{-1.7} $\downarrow$ \\
\midrule
MaPLe \cite{maple}                                    & 84.6           & 91.8          & 7.2          &               & 83.6          & 94.2          & 10.6         &               \\
\textbf{Ours-MaPLe }      & \textbf{85.4}  & \textbf{87.4} & \textbf{2.0} & \cellcolor{LightRed} \textbf{-5.2} $\downarrow$ & \textbf{85.4} & \textbf{90.1} & \textbf{4.7} & \cellcolor{LightRed}\textbf{-5.8} $\downarrow$ \\
\midrule
PromptSRC \cite{promptsrc}                                & 85.7           & 92.1          & 6.4          &               & 89.4          & 93.4          & 3.9          &               \\
\textbf{Ours-PromptSRC }   & \textbf{89.6}  & \textbf{90.7} & \textbf{1.0} & \cellcolor{LightRed} \textbf{-5.4} $\downarrow$ & \textbf{90.2} & \textbf{91.1} & \textbf{0.9} & \cellcolor{LightRed} \textbf{-3.1} $\downarrow$ \\ \bottomrule
\end{tabular}
\caption{\textbf{Cross Robustness Evaluation on Waterbird and CelebA Datasets.} Here we applied our $\mathcal{L}_{\text{Aligned}}$ on the baseline method. "WG" refers to the Worst-Group accuracy and "Avg" refers to the Average-Group accuracy. The A downward arrow $\downarrow$ indicates a reduction in the accuracy gap for the groups. The decrease in average accuracy is normal in the group robustness setting, as the goal is to reduce the accuracy \textbf{gap} between groups. }
\label{waterbird}
\end{minipage}
\end{table}

\begin{table}[htb!]
\begin{minipage}[t]{0.45\textwidth}
\setlength{\tabcolsep}{5pt}
\fontsize{7}{7}\selectfont
\newcommand{\fivept}{\fontsize{4}{4}\selectfont\color{black}}
\newcommand{\sixept}{\fontsize{6}{5}\selectfont}
\newcommand{\redpt}{\fontsize{7}{8}\selectfont \color{blue}}
\begin{tabular}{@{}c|ccccccc@{}}
\toprule
                                   & Source        & \multicolumn{4}{c}{Target}                                      &               &              \\ 
\cmidrule(lr){2-2} \cmidrule(lr){3-7}

                                       &{ImageNet}      &{V2}   &{S} &{A}    &{R}    &{Avg}               \\
                                          \midrule
ProGrad \cite{prograd}                                  & 70.5          & 63.4          & 48.2            & 49.5          & 75.2          & 59.0                        \\
KgCoOp \cite{kgcoop}                                    & 71.2          & 64.1          & 49.0            & 50.7          & 76.7          & 60.1                       \\
CLIPOOD \cite{clipood}                                  & 71.6          & 64.9          & 49.3            & 50.4          & 77.2          & 60.5                       \\
HPT \cite{hpt}                                      & 71.7          & 65.3          & 49.4            & 50.9          & 77.4          & 60.7                     \\
CoPrompt \cite{coprompt}                                   & 70.8          & 64.3          & 49.4            & 50.5          & 77.5          & 60.4                       \\
ArGue-N \cite{argue}                                      & 71.8          & 65.0          & 49.3            & 51.5          & 77.0          & 60.7                        \\
\toprule
LPT \cite{maple}                                      & 71.7 & 64.4 & 48.8 & 50.5 & 75.5 & 59.8 \\
\textbf{Ours-LPT}        & \textbf{71.9} & \textbf{64.7} & \textbf{48.9} & \textbf{51.1} & \textbf{76.1} & \textbf{60.2} \\
\rowcolor{LightGray}
$\bigtriangleup$                          & -0.2 & +0.3 & +0.1 & +0.6 &\cellcolor{LightRed} +0.6 & \cellcolor{LightRed}+0.4  \\
\midrule
VPT \cite{visual-prompt}                                      & 69.9 & 63.1 & 47.9 & 43.0 & 75.9 & 57.5 \\
\textbf{Ours-VPT}          & \textbf{70.2} & \textbf{63.3} & \textbf{48.5} & \textbf{45.1} & \textbf{76.0} & \textbf{58.2} \\
\rowcolor{LightGray}
$\bigtriangleup$                          & +0.3  & +0.2 & +0.6 &\cellcolor{LightRed} +2.1 & +0.1 & \cellcolor{LightRed}+0.8  \\
\midrule
MaPLe \cite{maple}                                    & 70.2 & 63.9 & 48.6 & 50.3 & 76.9 & 59.9 \\
\textbf{Ours-MaPLe }      & \textbf{70.8} & \textbf{64.1} & \textbf{51.2} & \textbf{51.7} & \textbf{77.8} & \textbf{61.2} \\
\rowcolor{LightGray}
$\bigtriangleup$                          & -0.6 & +0.2 &\cellcolor{LightRed} +2.6 & +1.3 & +0.9 & \cellcolor{LightRed}+1.3  \\
\midrule
PromptSRC \cite{promptsrc}                                & 71.3 & 64.2 & 49.2 & 50.8 & 77.7 & 60.5 \\
\textbf{Ours-PromptSRC  }   &  \textbf{71.5} & \textbf{65.4} & \textbf{51.9} & \textbf{52.0} & \textbf{77.8} & \textbf{61.8} \\
\rowcolor{LightGray}
$\bigtriangleup$                          & -0.3 & +1.1 & \cellcolor{LightRed}+2.7 & +1.2 & +0.1 & \cellcolor{LightRed}+1.3 \\ \bottomrule
\end{tabular}
% \captionsetup{width=0.9\textwidth}
\caption{\textbf{Out-of-Distribution Evaluation on ImageNet Variant Datasets.} Here ours method refers to applied both $\mathcal{L}_{\text{Aligned}}$ and $\mathcal{L}_{\text{MMD}}$ on the baseline method. In this setting, our method ensures the alignment of ImageNet data from different distributions, resulting in improved accuracy across all four datasets.}
\label{table:imagnet}    
\end{minipage}%
\end{table}

\begin{figure*}[htb!]
    \centering
    \includegraphics[width=\textwidth]{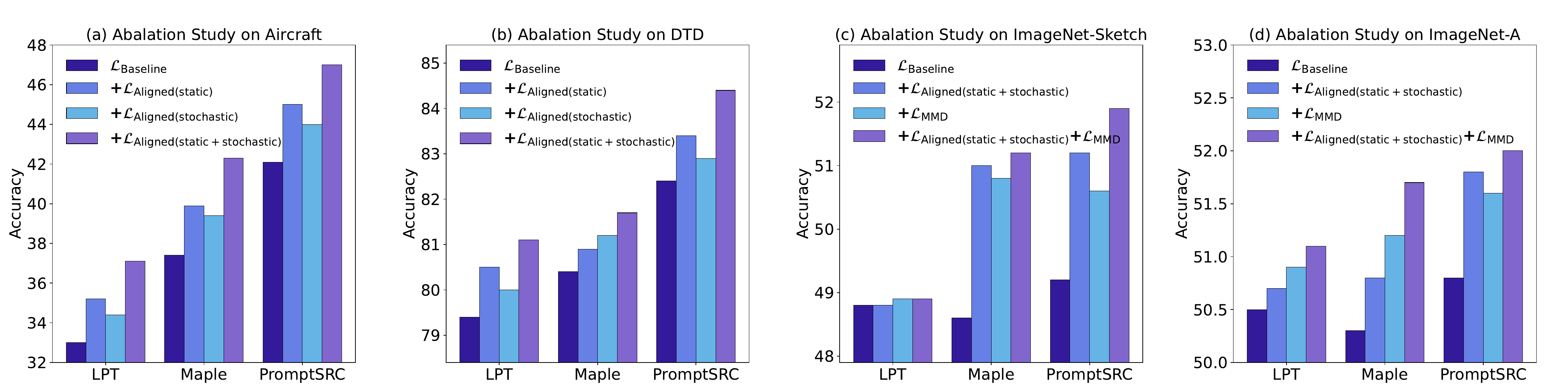}
    \caption{\textbf{The ablation study on individual datasets.} $\mathcal{L}_{\text{Baseline}}$ refers to the use of text-based cross-entropy loss in the method. Figures (a) and (b) demonstrate that adding $\mathcal{L}_{\text{Aligned (Static)}}$ or $\mathcal{L}_{\text{Aligned (Stochastic)}}$ can complementarily improve accuracy in in-distribution tasks. Figures (c) and (d) show that adding $\mathcal{L}_{\text{MMD}}$ further enhances accuracy across out-of-distribution tasks.}
    \label{fig-ablation}
\end{figure*}

\subsection{Feature Alignment Improves the Group Robustness}
We further evaluated the model's Group Robustness \cite{zhang2022contrastive} by using the Waterbirds \cite{waterbird} and CelebA \cite{celeba} datasets. The Waterbirds dataset is divided into four categories based on background (land or water) and bird type (land bird or water bird). The CelebA dataset is also categorized into four groups by hair color (blond or non-blond) and gender (male or female). The model was trained across all groups, aiming to reduce the classification performance gap between the minority group and the overall groups. This setup tests how well the prompt tuning model handles the spurious correlations.

\textbf{The proposed feature alignment method improves the worst group accuracy and reduces the performance gap.} From Table~\ref{waterbird}, we can see that models incorporating the feature alignment $\mathcal{L}_{\text{Aligned}}$ consistently reduce the accuracy gap between the worst-group and average-group. Specifically, for the Waterbirds dataset, applying our method to MaPLe reduced the accuracy gap by $5.2$ points, and applying it to PromptSRC reduced the gap by $5.4$ points. Similar trends are observed for the CelebA dataset, where applying $\mathcal{L}_{\text{Aligned}}$ to the four prompt tuning structures consistently reduced the performance gap. These results demonstrate that our approach effectively enhances the model's group robustness. Furthermore, compared to other state-of-the-art methods, such as Contrastive-Adapter \cite{zhang2022contrastive}, our best model, $\mathcal{L}_{\text{Aligned}}$ applied to PromptSRC, reduced the performance gap from $5.7$ to $1.0$, showcasing its competitiveness.

\subsection{Feature Alignment and MMD Improves the Out-of-Distribution Recognition}
In this setting, the model was trained on the ImageNet dataset and evaluated on its four variants: ImageNetV2 \cite{imagenetv2}, ImageNetSketch \cite{imagenet-s}, ImageNet-A \cite{imagenet-a}, and ImageNet-R \cite{imagenet-r}. Table \ref{table:imagnet} shows the results. It is observed that the proposed robust prompt tuning method (indicated by $\mathcal{L}_{\text {Aligned}}$ + $\mathcal{L}_{\text {MMD}}$ ) consistently improves accuracy across different ImageNet variants (V2, S, A, R) when compared to the baseline prompt tuning method. Regarding the maximum gain on individual datasets, VPT improved by $2.1$ points on ImageNet-A, and PromptSRC by $2.7$ points on the ImageNet-Sketch dataset. These improvements illustrate that, compared to traditional cross-entropy methods, our approach narrows the distributional gap between ImageNet and its four variants by utilizing the cross-domain discrepancy match loss, thereby enhancing its out-of-distribution generalization capabilities on a larger scale. Further analysis of the reasons behind these accuracy enhancements will be conducted in the Ablation Study.

\begin{figure}[htb!]
    \centering
    \includegraphics[width=0.5\textwidth]{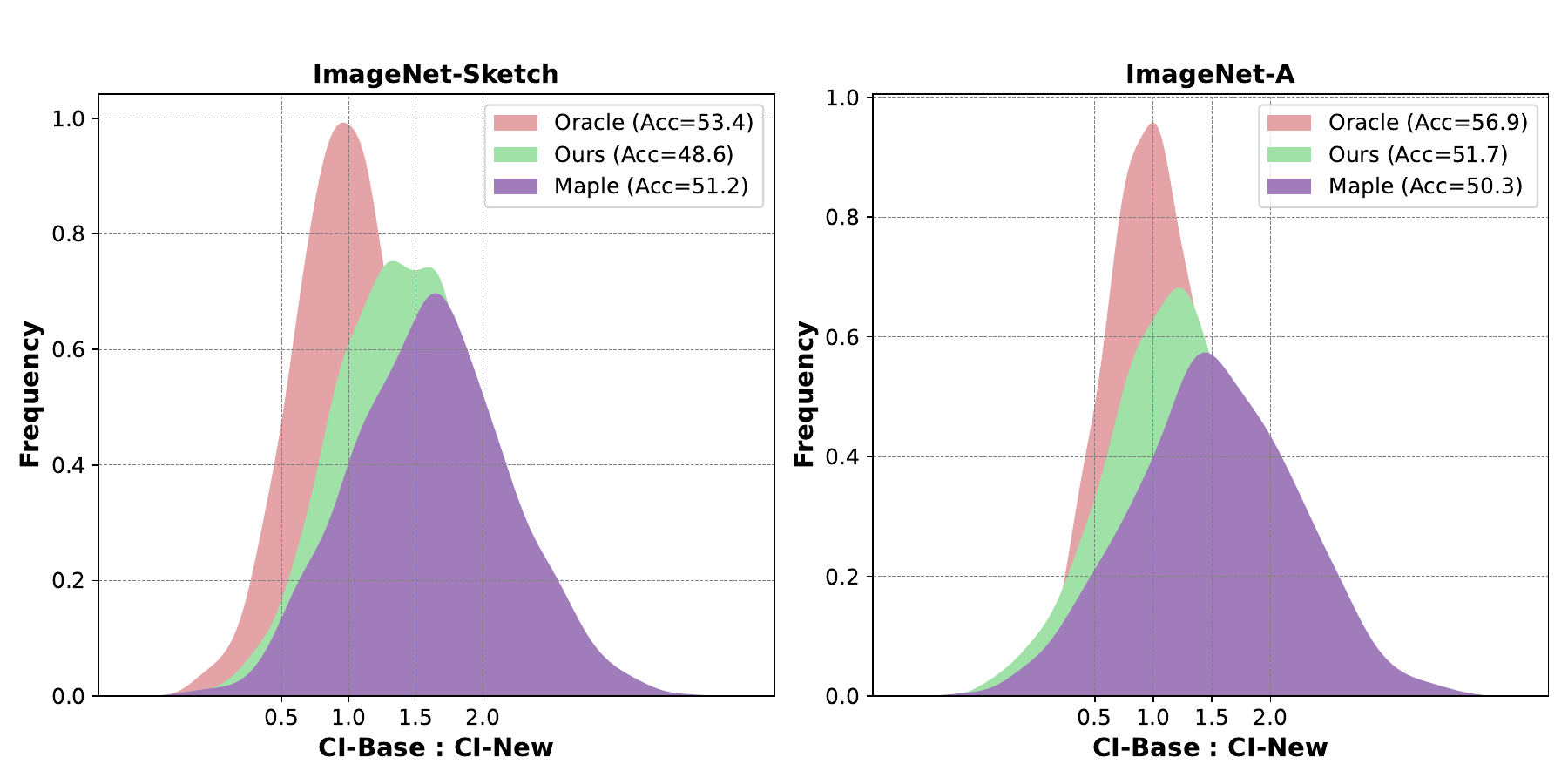}
    \caption{\textbf{The effectiveness analysis on channel importance ratio distribution. } The Oracle model is trained on the combination of labeled source and labeled target data, while our model is trained on the labeled source and unlabeled target data. Our $\mathcal{L}_{\text{MMD}}$ mitigate the domain shift of the baseline method and push the channel importance distribution close to the Oracle model.}
    \label{fig-mmd}
\end{figure}

\subsection{Abalation Studies}

\begin{table}[htb!]
\centering
\setlength{\tabcolsep}{2pt}
\fontsize{7}{7}\selectfont
\begin{tabular}{@{}ccccccc@{}}
\toprule
$\mathcal{L}_{\text{Baseline}}$ & $\mathcal{L}_{\text{Aligned (Static)}}$ & $\mathcal{L}_{\text{Aligned {(Stochastic)}}}$ & LPT  & VPT  & MaPLe & PromptSRC \\ \midrule
\checkmark                        &                                  &                                & 81.0 & 78.7 & 82.3  & 83.7      \\
\checkmark                        & \checkmark                        &                                & 82.0   & 79.9 & 83.7  & 84.8      \\
\checkmark                         &                                  & \checkmark                      & 81.6 & 79.1 & 82.6  & 84.0        \\
\checkmark                        & \checkmark                        & \checkmark                      & 82.2 & 80.5 & 84.2  & 85.5      \\ \bottomrule
\end{tabular}
\caption{\textbf{Ablation study of $\mathcal{L}_{\text{Aligned (Static)}}$ and $\mathcal{L}_{\text{Aligned {(Stochastic)}}}$ on Average Performance Across 11 Datasets.} $\mathcal{L}_{\text{Baseline}}$ refers to the use of text-based cross-entropy loss in the method. The results indicate that both $\mathcal{L}_{\text{Aligned (Static)}}$ and $\mathcal{L}_{\text{Aligned {(Stochastic)}}}$ complementarily improved the final accuracy.}
\label{tabel-ablation-align}
\end{table}

\begin{table}[htb!]
\centering
\setlength{\tabcolsep}{2pt}
\fontsize{7}{7}\selectfont
\begin{tabular}{cccccccc}
\toprule
$\mathcal{L}_{\text{Baseline}}$ & $\mathcal{L}_{\text{Aligned (Static)}}$ & $\mathcal{L}_{\text{Aligned {(Stochastic)}}}$ & $\mathcal{L}_{\text{MMD}}$ & LPT  & VPT  & MaPLe & PromptSRC \\ \midrule
\checkmark &                                                                 &                                                                  &                                 & 59.8 & 57.5 & 59.9  & 60.5      \\
\checkmark & \checkmark                                                       &                                                                  &                                 & 60.1 & 57.9 & 60.4  & 60.7      \\
\checkmark &                                                                 & \checkmark                                                        &                                 & 59.9 & 57.7 & 60.1  & 61.4      \\
\checkmark &                                                                 &                                                                  & \checkmark                       & 60.1 & 58.0 & 60.6  & 61.0      \\
\checkmark & \checkmark                                                       & \checkmark                                                        & \checkmark                       & 60.2 & 58.2 & 61.2  & 61.8      \\ \hline
\end{tabular}
\caption{ \textbf{Ablation study of $\mathcal{L}_{\text {MMD}}$ and $\mathcal{L}_{\text {Aligned}}$ on the out-of-distribution task.} The results show that $\mathcal{L}_{\text{MMD}}$ significantly enhances classification accuracy, while $\mathcal{L}_{\text{Aligned}}$ provides a slight improvement in accuracy.}
\label{table: ablation-ood}
\end{table}

\begin{figure*}[htb!]
\centering
\includegraphics[width=\textwidth]{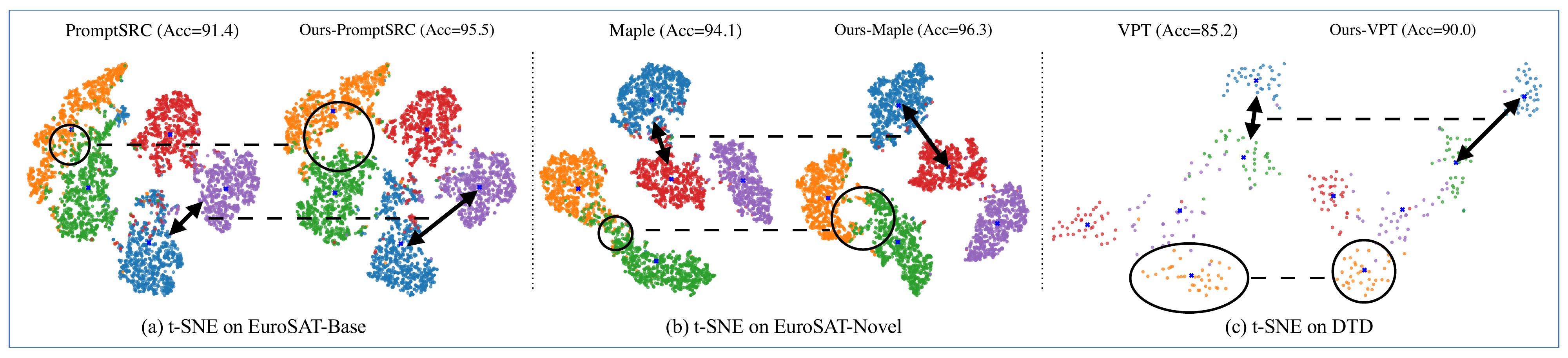}
\caption{\textbf{The t-SNE Visualization of Latent Embeddings.} The arrows in the three sub-figures illustrate our method can push the boundary between the two categories further apart. The circles in Figures (a) and (b) demonstrate that our method can separate the overlapping features of the two categories away from each other. The circle in Figure (c) shows that our method can achieve a more compact feature space.}
\label{fig:t-sne}
\end{figure*}

\textbf{Effect of Static and Stochastic Anchors.} We first conducted an ablation study on $\mathcal{L}_{\text{Aligned (Static)}}$ and $\mathcal{L}_{\text{Aligned (Stochastic)}}$, presenting the average accuracy across 11 datasets in Table \ref{tabel-ablation-align}. It was evident that combining the static anchor with the baseline method significantly enhanced the accuracy of the four prompt tuning models, with VPT increasing by $1.2$ points and MaPLe by $1.4$ points. We believe this enhancement was due to the extra supervision information introduced by the image anchor, which helped the model learn better decision boundaries. Furthermore, when combining the two losses, they complemented each other in enhancing model performance, resulting in a $1.9$-point improvement in VPT and a $2.0$-point improvement in MaPLe. We selected specific datasets and plotted bar charts of their ablation study results. As shown in Figure \ref{fig-ablation} (a) and (b), applying the Static Anchor or Stochastic Anchor separately with the baseline method led to an accuracy improvement on the Aircraft dataset, while the combination of all three losses yielded the optimal results.

\textbf{Effect of MMD}. Table \ref{table: ablation-ood} presents the out-of-distribution ablation experiment on ImageNet and its four variant datasets. We found that applying either the static anchor or stochastic anchor with the baseline method brings limited improvement on the OOD classification accuracy. In contrast, using $\mathcal{L}_{\mathrm{MMD}}$ resulted in significant improvements across the four different Prompt Tuning structures. When combining all these loss functions ($\mathcal{L}_{\mathrm{Aligned}}$ + $\mathcal{L}_{\mathrm{MMD}}$), we achieved the best enhancement in final classification performance. The ablation experiments on individual datasets, as shown in Figure \ref{fig-ablation} (c) and (d), demonstrated similar conclusions, indicating that the combination of all three losses jointly contributes to the improvement of OOD classification accuracy.

We further verify how our $\mathcal{L}_{\text {MMD }}$ addresses the domain shift. We train three different models for comparison: 1) the Oracle model, where the prompt is trained on the combination of Target and Source datasets; 2) the Baseline model, where only the text-based cross-entropy loss is used, and only the supervised Source data is used to train the prompt; and 3) our model, where our proposed method is used with both the supervised Source data and the unsupervised Target data. We use the metric of Channel Importance Ratio \cite{dept} to show the distribution shift. We conduct experiment on two datasets in Figures \ref{fig-mmd} (a) and \ref{fig-mmd} (b). The channel importance distribution shown in these figures demonstrates a clear gap between the Oracle model and the Baseline model. By employing our proposed $\mathcal{L}_{\text {MMD }}$ loss, which utilizes cross-domain discrepancy matching to overcome domain shift, we not only bring the Cl distribution close to the Oracle model but also increase recognition accuracy.

\begin{figure}[htb!]
\centering
\includegraphics[width=0.5\textwidth]{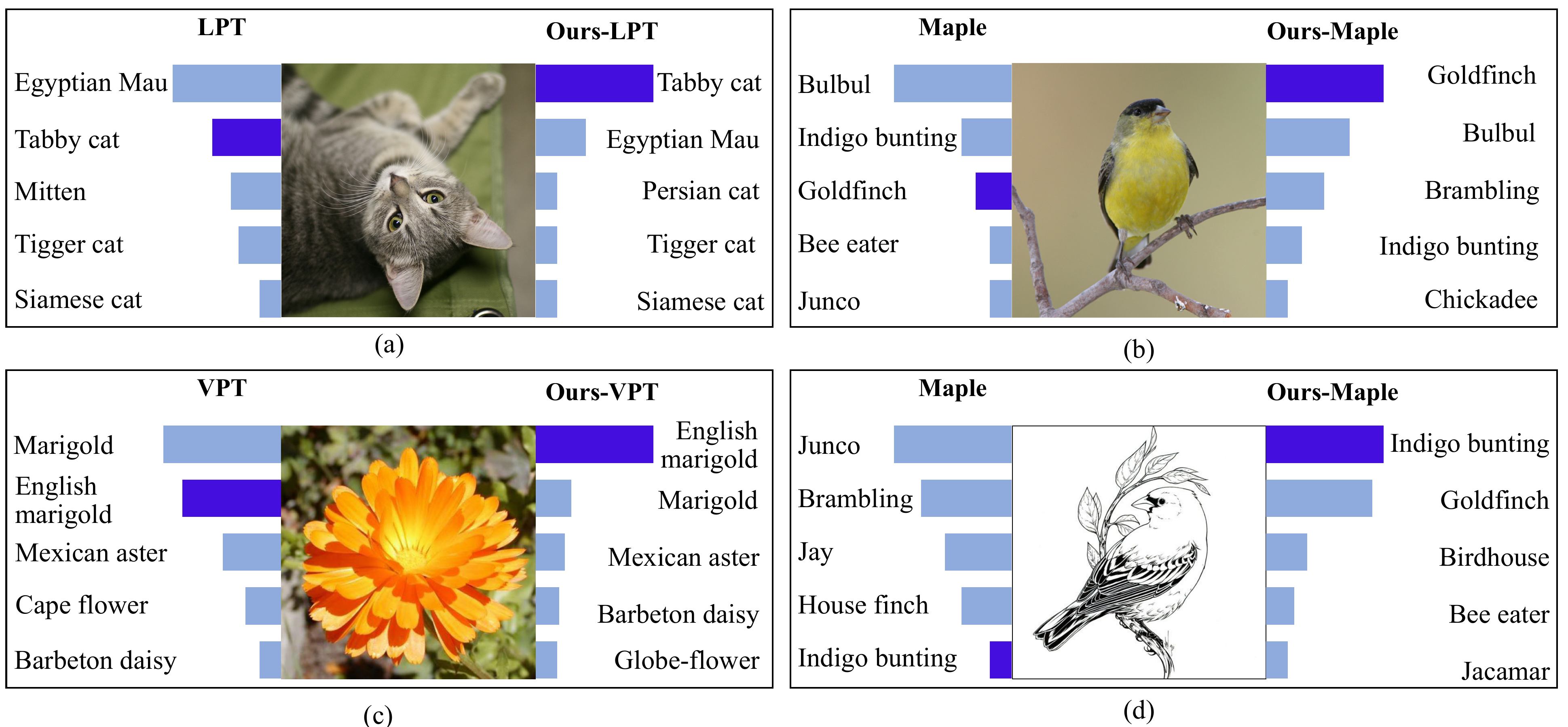}
\caption{\textbf{Comparison of Prediction Probabilities With and Without Our Method.} Our robust prompt tuning method effectively corrects misclassifications made by the baseline method.}
\label{fig-top5}
\end{figure}

\subsection{Aligned Feature Visualization}
Figure \ref{fig:t-sne} presents the results of image feature visualization. In Figure (a), it is evident that applying our $\mathcal{L}_{\text {Aligned}}$ method to MaPLe increases the distance between cluster centers of the same color in the visualized features. This indicates that our method enhances the learned latent space, bringing it closer to real samples, strengthening the model's decision boundaries, and consequently improving its accuracy. Similar improvements are observed in Figures \ref{fig:t-sne} (b) and (c). Additionally, Figure \ref{fig:t-sne} (c) demonstrates that our method achieves a more compact cluster representation, as highlighted by the circled areas. These results confirm the effectiveness of our $\mathcal{L}_{\text {Aligned}}$ method. %, which is designed based on relative representation.

Figure \ref{fig-top5} illustrates the qualitative results of our method on a single sample. In Figure \ref{fig-top5} (a), without using the $\mathcal{L}_{\text {Aligned}}$ loss function, an image of an Tabby cat was incorrectly classified as Egyptian Mau. However, after applying the $\mathcal{L}_{\text {Aligned}}$ loss function, the top-1 probability for this image corresponded to the correct category. Figures \ref{fig-top5}(b), (c) and (d) show similar results.
%, demonstrating that our proposed $\mathcal{L}_{\text {Aligned}}$ loss function improves the decision boundary, thereby correcting the misclassified class labels.

\section{Conclusion}
In this paper, we proposed a novel feature alignment method to mitigate the overfitting issue in prompt tuning. Our method combines static and stochastic anchors to learn a more aligned feature space. Based on this aligned space, we applied a modified cross-domain discrepancy matching loss to address domain shift. Experiments demonstrated that our approach outperforms existing methods in Base-to-Novel generalization, group robustness, and out-of-distribution tasks. We hope our work will inspire more research on the adaptation of visual language models.

\section{Acknowledgement} This work is partially supported by NSF AI Institute-2229873, NSF RI-2223292, an Amazon research award, and an Adobe gift fund. Prof. Lokhande thanks support provided by University at Buffalo Startup funds. We thank Prof. Won Hwa Kim (POSTECH South Korea) for the insightful discussions on the project. %Any opinions, findings and conclusions or recommendations expressed in this material are those of the author(s) and do not necessarily reflect the views of the National Science Foundation, the Institute of Education Sciences, or the U.S. Department of Education.

%%%%%%%%% REFERENCES
{\small
\bibliographystyle{ieee_fullname}
\bibliography{egbib}
}

\newpage
\appendix
\section{Overview of Appendix}

In addition to the main method and experiments outlined in the paper, we offer supplementary information about our work in the Appendix. In Appendix B, we delve into implementation within the methodology section. We provide more details about our implementation in the Static Anchor and Stochastic Anchor, as well as the Maximum Mean Discrepancy Minimization. Subsequently in Appendix C, we provide more results on dataset setting and adaptation experiments, covering the Base-to-Novel Generalization and group Robustness task. Further in Appendix D, we present additional results of ablation studies along with their visualization outcomes. Lastly in Appendix E and F, we explore the limitations of our work and analyze its broader impact.

\section{Method}\label{sec:method} 
\subsection{Review the Adaption of CLIP}
We first review the pretraining and inference stage of the CLIP model, then we discuss the adaptation of CLIP. During the pretraining phase, a large-scale dataset of image-text pairs ${(x,y)}$ is collected for training the model using a contrastive learning approach. Here $x$ represents an image, and $y$ denotes its corresponding textual description. For each image $x$, an image encoder model $f_{\theta}$ is parameterized by ${\theta}$ to extract its visual feature vector ${u} \in \mathbb{R}^{1 \times H}$: ${u}=f_{\theta}(x)$. Similarly, for each textual description ${y}$, a text encoder $g_{\phi}$ is parameterized by ${\phi}$ to get its feature embedding ${v} \in \mathbb{R}^{1 \times H}$: ${v}=g_{\phi}({y})$. For the $i$-th image $x_i$ and the $j$-th language description ${y}_j$ in a batch $\mathcal{B}$, we normalize their feature vectors to a hyper-sphere using:${u}_i=\frac{f_{\theta}\left(x_i\right)}{\left\|f_{\theta}\left(x_i\right)\right\|}$ and ${v}_j=\frac{g_{\phi}\left({y}_j\right)}{\left\|g_{\phi}\left({y}_j\right)\right\|}$.

\textbf{Test phase of CLIP.} In this phase, a predefined prompt "a photo of a {}"  is commonly employed for inference. Let's consider a single test image $x_{\text {test }}$ of class $C$, where $x_{\text {test}} \in \mathbb{R}^{C \times H \times W}$ and $C \in \mathbb{R}^K$ for a $K$-class classification problem. The predefined prompt is prepended to each class label in $C$ to construct the language description. The zero-shot prediction probability for the test image is determined by:

\begin{equation}
Pr\left( c=k \mid x_{\text {test }}\right)=\frac{\exp \left(\operatorname{sim}\left({u}, {v}_{{i}}\right) \tau\right)}{\sum_{i=1}^K \exp \left(\operatorname{sim}\left({u}, {v}_{{i}}\right) \tau\right)}
\label{test}
\end{equation}

\subsection{Introduction of the Prompt Tuning Method}
\textbf{Language Prompt Tuning} involves introducing learnable parameters into the text branch. We follow the same notation in \cite{maple} and \cite{self-maple}. In the text branch, the class label $c$ is formatted as a language description within a text template as ``a photo of a \{label\}", which can be further transferred as $\tilde{{y}}=$ $\left\{{t}_{S O S}, {t}_1, {t}_2, \cdots, {t}_L, {c}_n, {t}_{E O S}\right\}$. Here ${{t}_l}$ are the word embeddings of the text template, and ${c}_n$ are the class label. The ${t}_{S O S}$ and ${t}_{E O S}$ are the learnable start and end token embeddings. The text encoder $g$ encodes the input tokens $\tilde{{y}}$ through multiple transformer blocks to generate a latent text feature representation $\tilde{{g}}=g\left(\tilde{{y}}, \theta_g\right)$. In Language prompt tuning, learnable text prompts $\Lambda_\text{text} \in \left\{\lambda_{\text{text}}^1, \lambda_{\text{text}}^2, \cdots, \lambda_{\text{text}}^L\right\}$
are appended to the input ${\tilde{{y}}}$. In CoOp \cite{coop} or CoCoOp \cite{CoCoOp}, the learnable text prompts ${\mathbb \lambda}_{{text}}$ are only added to input of text encoder. While in Maple \cite{maple}, the learnable text prompts are appended to multiple transformer layers as $\left[\ldots, \tilde{{y}}_i\right]=\mathcal{L}_i\left(\left[\Lambda_\text{text}, \tilde{{y}}_{i-1}\right]\right) \quad i=1,2, \cdots, J$, where $\mathcal{L}_i$ represent the $i$ layer number in the transformer, $J$ represent the prompt depth.

\textbf{Visual Prompt Learning} involves the integration of learnable prompts within the image branch. The input image $x$ is divided into $M$ patches, and these patches are projected to generate patch embeddings $\tilde{x}=$ $\left\{{e}_{c l s}, {e}_1, {e}_2, \cdots, {e}_M\right\}$, where ${e}_{c l s}$ is the learnable class token. Subsequently, learnable visual prompts are introduced as  $\Lambda_\text{visual} \in \left\{\lambda_{\text{visual}}^1, \lambda_{\text{visual}}^2, \cdots, \lambda_{\text{visual}}^L\right\}$
. The learnable visual prompts are appended to multiple transformer layers as $\left[c_i, \tilde{x}_i, \ldots\right]=\mathcal{V}_i\left(\left[c_{i-1}, \tilde{x}_{i-1}, \Lambda_\text{visual}\right]\right) \quad i=1,2, \cdots, J$, where $\mathcal{V}_i$ represent the $i$ layer in vision transformer, $J$ represent the prompt depth. 

\textbf{Multi-modal Prompt Learning} integrates language prompt learning and visual prompt learning, combining them synergistically. Simply combining text prompt learning and visual prompt learning is called independent prompt learning, which is used in UPT \cite{upt}. To foster interaction between the image and text branches, multi-modal prompt learning employs projection layers ${L}_{{t}}=\left\{\tilde{{l^1}}, \tilde{{l^2}}, \cdots, \tilde{{l^J}}\right\}$ for projecting the learnable language prompts onto the visual prompts, defined as $\Lambda_\text{visual}=\left\{\tilde{{l^1}}(\lambda_{\text{text}}^1), \tilde{{l^2}}(\lambda_{\text{text}}^2), \cdots, \tilde{{l^J}}(\lambda_{\text{text}}^L)\right\}$. This formulation facilitates interaction between the visual and language prompts. Such unified prompt tuning is a key feature of the Maple \cite{maple} framework.

% \subsection{Existing Text-Based Cross Entropy Loss.} Consider an image classification task where a batch of images ${(x_n})_{n=1}^B$ and class labels ${({c}_i})_{i=1}^K$ are provided from the source dataset. Tokenized image embeddings and learnable visual prompts are fed into the image encoder, resulting in image features ${u}=f_{{\theta}}\left({X, \Lambda_\text{visual}} \right), {u} \in \mathbb{R}^{B \times H}$, $H$ denotes the hidden dimension in the CLIP model. Similarly, for the text encoder, tokenized class labels and learnable text prompts are processed, yielding text embeddings ${v}=g_{{\phi}}\left(\mathbb{A}_{\text Y}, { \Lambda_\text{text}}\right), {v} \in \mathbb{R}^{K \times H}$. As ${{\mathbb{A}}}_{\text Y}$ is fixed during training, all image features are trained to align with these text embeddings, thus we call them \textbf{text anchors}. The dimension of the final logits between the image feature and text anchors is $[B, K]$.  The text anchor-based cross-entropy loss is given below:
% \begin{equation}
%  \mathcal{L}_{\text{CE}}=-\frac{1}{B} \sum_{j=1}^B \frac{\exp \left(\operatorname{sim}\left({u}_{{j}}, v_{i}\right) \tau\right)}{\sum_{i=1}^K \exp \left(\operatorname{sim}\left({u}_{{j}}, v_{i}^{k}\right) \tau\right)}  \cdot \mathbb{I}_{[y_{j}=c_{i}]}
%  \label{l-ce}
% \end{equation}

\subsection{Static Anchor Implementation}

To address the overfitting issues of text-based cross-entropy loss in scenarios with limited data, we propose a symmetrical static anchor alignment method, analogous to an image-based cross-entropy loss. This method involves two primary steps:

\textbf{Step 1: Construction of Image Anchors.} We use a pre-trained CLIP image encoder to extract features for each category in the source dataset, followed by K-means clustering to identify the centroid of each category's features, denoted as ${a_{x}^{k}}$. It is important to note that the dimensionality of ${a_{x}^{k}}$ differs from that of the batch image features $f_{\theta}(x)$.

\textbf{Step 2: Alignment with Text Samples.} For each image batch, corresponding text labels represented as language descriptions are aligned, with batch text embeddings ${v'} = g_{\phi}(y, \Lambda_{\text{txt}})$, where ${v'} \in \mathbb{R}^{B \times H}$, also differing in feature dimensions from class labels.

\subsection{Stochastic Anchors Implementation}

Stochastic anchors, selected during each batch, can be implemented as cross-modal contrastive learning process. Traditional supervised learning, which models relationships between images and discrete labels, often neglects textual concepts associated with labels. In contrast, stochastic anchors learning fosters understanding of visual concepts through enforcing batch-level text-image alignment.

We construct a contrastive similarity matrix ${s'} = \operatorname{sim}(u, v')$, where ${s'} \in \mathbb{R}^{N \times N}$. This matrix supports the formulation of both image-to-text and text-to-image contrastive losses, averaging these to derive the final text-image contrastive loss. In this matrix, only diagonal entries are treated as positive examples, enhancing the robustness of the latent space by introducing a larger set of negative samples.

\subsection{Maximum Mean Discrepancy Implementation}

Maximum Mean Discrepancy (MMD) is a kernel-based method primarily used to test the equality of two distributions from samples. Introduced in \cite{gretton2012kernel}, MMD compares the mean embeddings in a feature space, facilitating its use as a loss function in various machine learning tasks, including density estimation, generative modeling, and inverse problems tackled with invertible neural networks. Its simplicity and robust theoretical foundations make MMD particularly advantageous.

\blue{
To compute MMD for multi-modal data, a product measure is constructed to create a new probability space. Combining two probability spaces increases the complexity of the resulting $\sigma$-algebra, necessitating additional samples to characterize the probability space, as noted in Fact 3. We propose Equation 5 to define an induced measure, specifically the anchor-aligned probability measure, as a replacement for the traditional product measure in MMD computation. Equation 5 is essential for the application of MMD in anchor-aligned feature spaces. The transformation of the original probability measure $P_x$ via an anchor-aligned mapping is demonstrated. This equation defines a new probability measure, $P_x^{a_y}$, corresponding to the anchor $a_y$. }

\blue{Equation 6 specifies the MMD loss between two distributions, $\P_{{x}}^{id}$ (in-domain) and $\P_{{x}}^{ood}$ (out-of-domain), within the anchor-aligned feature space. This equation quantifies the discrepancy between two probability distributions in the anchor-aligned feature space. Here is how to use Equation 6 for the current task: The first term computes the expectation over all $x_{\text {id }}$ samples, while the second term computes the expectation over all $x_{\text {ood }}$ samples. Here, $k$ is the kernel function, and in our experiments, we use the Gaussian kernel. $f$ is the image encoder, and $\theta$ is being updated during training. In practical implementations, the expectations are replaced with sample averages.
}

% We define MMD using two equivalent forms: as a distance between feature means and as an integral probability metric. Given random variables $X$ from distribution $P$ and $Y$ from distribution $Q$, the feature map $\beta$ projects $X$ and $Y$ to a new space $\mathcal{\beta}$. Using the kernel trick, the inner product in $\mathcal{\beta}$ is defined as:
% \begin{equation}
%     k(X, Y) = \langle\beta(X), \beta(Y)\rangle_F
% \end{equation}

% Feature means, or mean embeddings, compute the coordinate-wise means of these projections:
% \begin{equation}
%     \mu_P(\beta(X)) = [E[\beta(X_1)], \dots, E[\beta(X_m)]]^T
% \end{equation}

% The inner product of feature means is then expressed in terms of the kernel function:
% \begin{equation}
%     \langle \mu_P(\beta(X)), \mu_Q(\beta(Y)) \rangle_F = E_{P, Q}[k(X, Y)]
% \end{equation}

% The squared MMD value, estimating the distributional discrepancy, is computed as:
% \begin{equation}
%     MMD^2(P, Q) = \|\mu_P - \mu_Q\|_F^2
% \end{equation}

% In practical applications, for instance, aligning image features distribution from the source domain to the target domain, MMD helps minimize the discrepancy between cross-domain distributions. This is critical in tasks like domain adaptation, where source domain logit $s^s = \operatorname{sim}(u, v)$ and target domain logit $s^t = \operatorname{sim}(u^t, v)$ are matched.

\section{Experiments Protocol}
\subsection{Base-to-Novel Dataset Split}

In our experiment, we partition all class samples into two distinct groups, as outlined in the tables: the Base group (Table \ref{base}) and the Novel group (Table \ref{novel}).

Consider the ImageNet dataset illustrated in Table \ref{all}, which consists of 1,000 classes. We divide the training set into two subsets, each containing 500 non-overlapping classes. For instance, one subset may include classes such as ["tench", "goldfish", "great white shark", "tiger shark", ...], and the other might feature ["spindle", "sports car", "spotlight", ...]. This separation ensures that no class from one group appears in the other, thereby preventing the model from encountering unknown classes during training and enhancing the fairness and credibility of our out-of-distribution evaluations. The test set follows a similar bifurcation, maintaining correspondence with the class labels from the training set.

\begin{table}[htp!]
\centering
\setlength{\tabcolsep}{1.5pt}
\fontsize{7}{8}\selectfont
\newcommand{\fivept}{\fontsize{5}{4}\selectfont\color{black}}
\newcommand{\sixept}{\fontsize{6}{5}\selectfont}
\newcommand{\redpt}{\fontsize{7}{8}\selectfont \color{blue}}

\begin{tabular}{@{}lllll@{}}
\toprule
             & Classes & Train-Samples & Val-Samples & Test-Samples \\ \midrule
OxfordPets   & 18      & 288           & 368         & 1881         \\
Flowers102   & 51      & 816           & 817         & 1053         \\
FGVCAircraft & 50      & 800           & 1667        & 1666         \\
DTD          & 23      & 368           & 564         & 864          \\
EuroSAT      & 5       & 80            & 2700        & 4200         \\
StanfordCars & 98      & 1568          & 818         & 4002         \\
Food101      & 50      & 800           & 10100       & 15300        \\
SUN397       & 198     & 3168          & 1985        & 9950         \\
Caltech101   & 50      & 800           & 825         & 1549         \\
UCF101       & 50      & 800           & 949         & 1934         \\
ImageNet     & 500     & 8000          & 25000       & 25000        \\ \bottomrule
\end{tabular}

\caption{\textbf{Base class samples statistics.} The first column "Classes" denotes the total number of classes for each category. The columns "Train-Samples", "Val-Samples", and "Test-Samples" represent the respective number of images allocated for model training, validation, and testing purposes. }
\label{base}
\end{table}

\begin{table}[htp!]
\centering
\setlength{\tabcolsep}{1.5pt}
\fontsize{7}{8}\selectfont
\newcommand{\fivept}{\fontsize{5}{4}\selectfont\color{black}}
\newcommand{\sixept}{\fontsize{6}{5}\selectfont}
\newcommand{\redpt}{\fontsize{7}{8}\selectfont \color{blue}}
\begin{tabular}{@{}lllll@{}}
\toprule
             & Classes & Train-Samples & Val-Samples & Test-Samples \\ \midrule
OxfordPets   & 19      & 304           & 368         & 1788         \\
Flowers102   & 51      & 816           & 816         & 1410         \\
FGVCAircraft & 50      & 800           & 1,666       & 1667         \\
DTD          & 24      & 384           & 564         & 828          \\
EuroSAT      & 5       & 80            & 2,700       & 3900         \\
StanfordCars & 98      & 1568          & 817         & 4039         \\
Food101      & 51      & 816           & 10,100      & 15000        \\
SUN397       & 199     & 3184          & 1,985       & 9900         \\
Caltech101   & 50      & 800           & 824         & 916          \\
UCF101       & 51      & 816           & 949         & 1849         \\
ImageNet     & 500     & 8000          & 25000       & 25000        \\ \bottomrule
\end{tabular}
\caption{\textbf{Novel class samples statistics.} The first column "Classes" denotes the total number of classes for each category. The columns "Train-Samples", "Val-Samples", and "Test-Samples" represent the respective number of images allocated for model training, validation, and testing purposes.}
\label{novel}
\end{table}

\begin{table}[htp!]
\centering
\setlength{\tabcolsep}{1.5pt}
\fontsize{7}{8}\selectfont
\newcommand{\fivept}{\fontsize{5}{4}\selectfont\color{black}}
\newcommand{\sixept}{\fontsize{6}{5}\selectfont}
\newcommand{\redpt}{\fontsize{7}{8}\selectfont \color{blue}}
\begin{tabular}{@{}llllll@{}}
\toprule
             & Classes & Train-Samples & Val-Samples & Test-Samples & Task             \\ \midrule
OxfordPets   & 37      & 2944          & 736         & 3669         & Fine-Grained     \\
Flowers102   & 102     & 4093          & 1633        & 2463         & Fine-Grained     \\
FGVCAircraft & 100     & 3334          & 3333        & 3333         & Fine-Grained     \\
DTD          & 47      & 2820          & 1128        & 1692         & Textures         \\
EuroSAT      & 10      & 13500         & 5400        & 8100         & Satellite Images \\
StanfordCars & 196     & 6509          & 1635        & 8041         & Fine-Grained     \\
Food101      & 101     & 50500         & 20200       & 30300        & Food             \\
SUN397       & 397     & 15880         & 3970        & 19850        & Scene            \\
Caltech101   & 100     & 4128          & 1649        & 2465         & Object           \\
UCF101       & 101     & 7639          & 1898        & 3783         & Action           \\
ImageNet     & 1000    & 12800000      & N/A         & 50000        & Object           \\ \bottomrule
\end{tabular}
\caption{ \textbf{All class samples statistics from the original datasets.} The last column "task" provides a broad categorization of these image classification tasks, such as fine-grained classification or texture classification.}
\label{all}
\end{table}

\subsection{Group Robustness Baseline}

For the group robustness experiment described in Section 4.4, we give a more comprehensive introduction about the baseline method that we compared with.

We evaluate our method against several methods in group robustness experiments, including zero-shot classification, ERM linear probing \cite{kumar2022fine}, and ERM adapter training \cite{clip-adapter}. Additionally, we compare against recent approaches tailored to enhancing downstream transfer in analogous scenarios, all while utilizing only pretrained model embeddings \cite{adila2023zero}.

One such method is Weight space ensembling (WiSE-FT) \cite{wortsman2022robust}, which initially trains a linear classifier using standard ERM and then combines the classifier outputs with the initial zero-shot predictions. Although originally proposed for training linear classifiers and fine-tuning the original weights of a foundational model, we focus on the prompt tuning in the extra parameter in our context.

Another approach is Deep feature reweighting (DFR) \cite{kirichenko2022last}, which entails training a linear probe on embeddings computed from a pretrained model over group-balanced data. Similar to previous studies \cite{liu2021just, zhang2022correct}, we treat incorrectly and correctly classified samples as proxies for distinct groups.

Lastly, we consider the Contrastive Adapter approach \cite{zhang2022contrastive}, which introduces contrastive adapting. This method trains adapters with contrastive learning to bring sample embeddings closer to both their ground-truth class embeddings and other sample embeddings within the same class. While our method differs from this work, as we apply Contrastive learning to Prompt Tuning instead of Adapters.

\subsection{Training Details}
We utilized SGD as the optimizer optimizer with an initial learning rate of 0.0025 for Batch size 4, and a learning rate of 0.01 for batch size 128. The cosine annealing strategy is chosen to schedule the learning rate. For the Base to Novel Generalization setting, we use a few-shot training of 16 shots with a training duration of 20 epochs, while for Group Roubustness,  we train 10 epochs on Waterbird and 5peochs on CelebA for the full dataset. All images were resized to 224x224 pixels, utilizing the same image preprocessing technique for the CLIP image encoder. All CLIP models adopted the ViT-B/16 backbone. We maintained consistency across all other settings as the baseline work, making modifications solely to the loss function to ensure a fair comparison between our method and the standard cross-entropy loss.

\section{More Experiments Results}
\blue{\subsection{Anchor Selections Comparison}}
\blue{To evaluate whether different static anchor selections affect the final results, we conducted the ablation study on the anchor selection experiment, with the results shown in Table \ref{k-means}. We used the pre-trained CLIP model with a ViT-B/16 backbone as the feature extractor. All training images from each dataset were fed into the model's image encoder, and the resulting features were stored. The features are grouped by the ground truth label, then we use different anchor selection methods to choose the most representative one as the static anchor. The anchor selection methods we have are (1) K-means: we utilize the cluster center of K-means as the static anchor; (2) Hierarchical clustering: also the cluster center is utilized as the static anchor; (3) Group means, we direct calculation of the mean features for all the samples in each group. Table \ref{k-means} shows that K-means and other methods do not have significant differences, while K-means yield better results compared to the hierarchical clustering method.}

% % Please add the following required packages to your document preamble:
% % \usepackage{booktabs}
% % \usepackage{multirow}
% \blue{\begin{table}[]
% \centering
% \setlength{\tabcolsep}{1.2pt}
% \fontsize{7}{8}\selectfont
% \newcommand{\fivept}{\fontsize{5}{4}\selectfont\color{black}}
% \newcommand{\sixept}{\fontsize{6}{5}\selectfont}
% \newcommand{\redpt}{\fontsize{7}{8}\selectfont \color{blue}}
% \begin{tabular}{@{}l|llllllllll|l@{}}
% \toprule
%                       & Method  & Pets & Flowers & Aircraft & DTD  & EuroSAT & Cars & Food & Caltech & UCF  & Avg \\ \midrule
% \multirow{2}{*}{Base}  & G-Means & 95.4 & 97.6    & 43.2     & 82.9 & 91.9    & 78.6 & 90.7 & 98.2    & 86.9 & 85.0    \\
%                        & K-Means & 95.3 & 97.5    & 43.0     & 83.3 & 92.4    & 78.8 & 90.6 & 98.1    & 86.5 & \textbf{85.1}    \\
%                        \midrule
% \multirow{2}{*}{Novel} & G-Means & 97.5 & 77.2    & 38.2     & 63.6 & 68.1    & 75.4 & 91.7 & 94.5    & 78.2 & 76.0    \\
%                        & K-Means & 97.5 & 77.7    & 36.9     & 63.9 & 79.4    & 75.2 & 91.7 & 94.1    & 79.1 & \textbf{77.3}    \\ %\cmidrule(l){2-12} 
% \bottomrule
% \end{tabular}
% \caption{ \blue{\textbf{K-means Anchor Selection Vs Group-Means Anchor Selection.} The 'Avg' represents the average accuracy over all the datasets.}}
% \label{k-means}
% \end{table}}

% \usepackage{booktabs}
% \usepackage{multirow}
\blue{\begin{table}[]
\centering
\setlength{\tabcolsep}{1.2pt}
\fontsize{7}{8}\selectfont
\newcommand{\fivept}{\fontsize{5}{4}\selectfont\color{black}}
\newcommand{\sixept}{\fontsize{6}{5}\selectfont}
\newcommand{\redpt}{\fontsize{7}{8}\selectfont \color{blue}}
\begin{tabular}{@{}l|c|ccccccccc|l@{}}
\toprule
                      & Method  & Pets & Flowers & Aircraft & DTD  & EuroSAT & Cars & Food & Caltech & UCF  & Avg \\ \midrule
\multirow{3}{*}{Base}  & G-Means & 95.4 & 97.6    & 43.2     & 82.9 & 91.9    & 78.6 & 90.7 & 98.2    & 86.9 & 85.0    \\
& H-Cluster & 95.2 & 97.5    & 43.1     & 83.0 & 92.2    & 78.7 & 90.5 & 98.0    & 86.7                      & 85.0          \\
                       & K-Means & 95.3 & 97.5    & 43.0     & 83.3 & 92.4    & 78.8 & 90.6 & 98.1    & 86.5 & \textbf{85.1}    \\
                       \midrule
\multirow{3}{*}{Novel} & G-Means & 97.5 & 77.2    & 38.2     & 63.6 & 68.1    & 75.4 & 91.7 & 94.5    & 78.2 & 76.0    \\
& H-Cluster & 97.4 & 77.1 & 36.7 & 64.1 & 72.5 & 74.9 & 90.5 & 94.2 & 78.5 & 76.2 \\
                       & K-Means & 97.5 & 77.7    & 36.9     & 63.9 & 79.4    & 75.2 & 91.7 & 94.1    & 79.1 & \textbf{77.3}    \\ %\cmidrule(l){2-12} 
\bottomrule
\end{tabular}
\caption{ \blue{\textbf{Anchor Selection Method Comparison.} K-means is the default anchor selection method used in this paper. G-Means represents the group means anchor selection method. H-Cluster means hierarchical clustering anchor selection method. The 'Avg' represents the average accuracy over all the datasets.}}
\label{k-means}
\end{table}}

\begin{figure*}[htb!]
  \centering
    \centering
    \includegraphics[width=\textwidth]{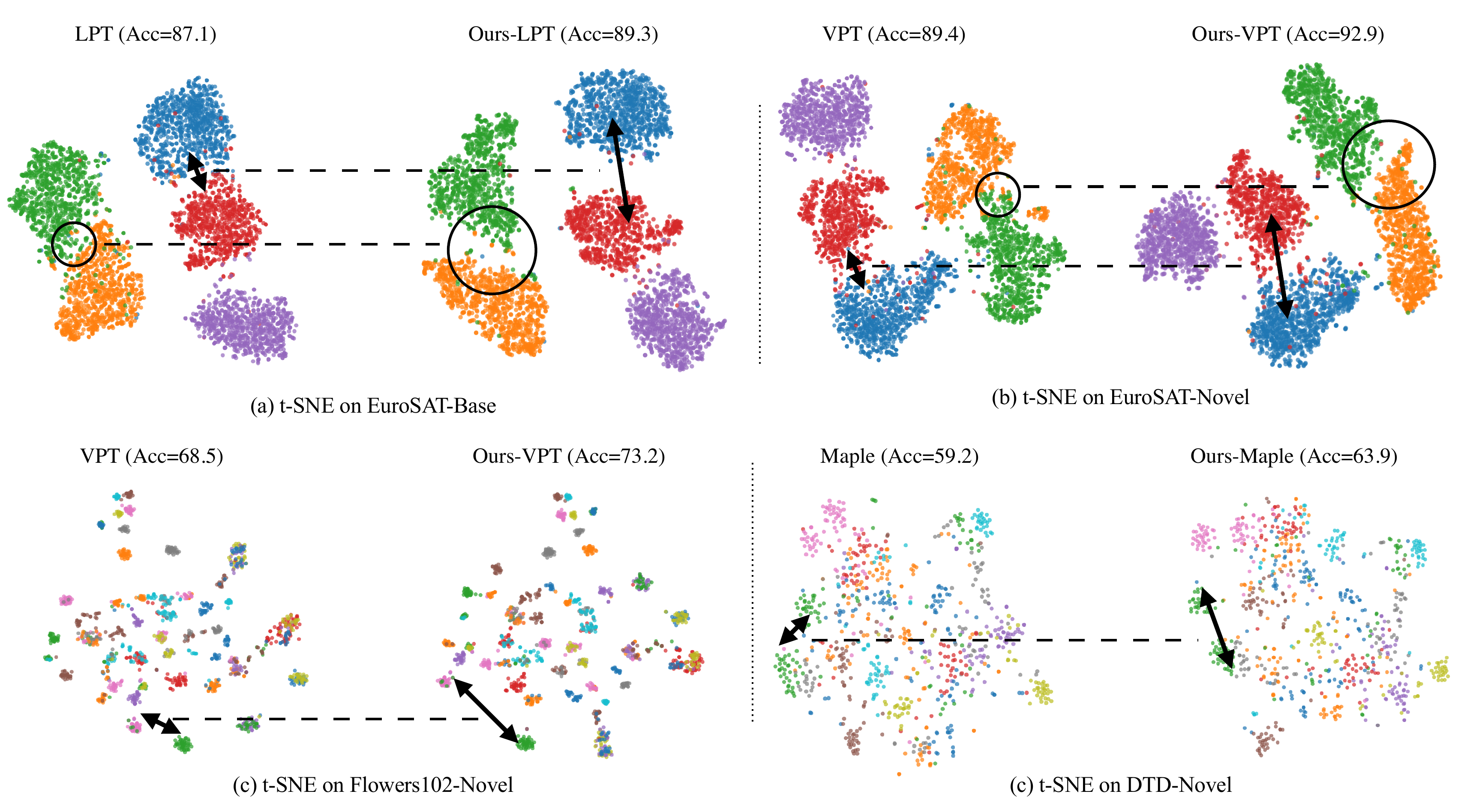}
  \caption{\textbf{The t-SNE Visualization of Latent Embeddings.} The arrows in the figures illustrate our method can push the boundary between the two categories further apart. The circles in Figures (a) and (b) demonstrate that our method can separate the overlapping features of the two categories away from each other.}
  \label{fig:t-sne}
\end{figure*}

\begin{figure*}[htb!]
    \centering
    \includegraphics[width=\textwidth]{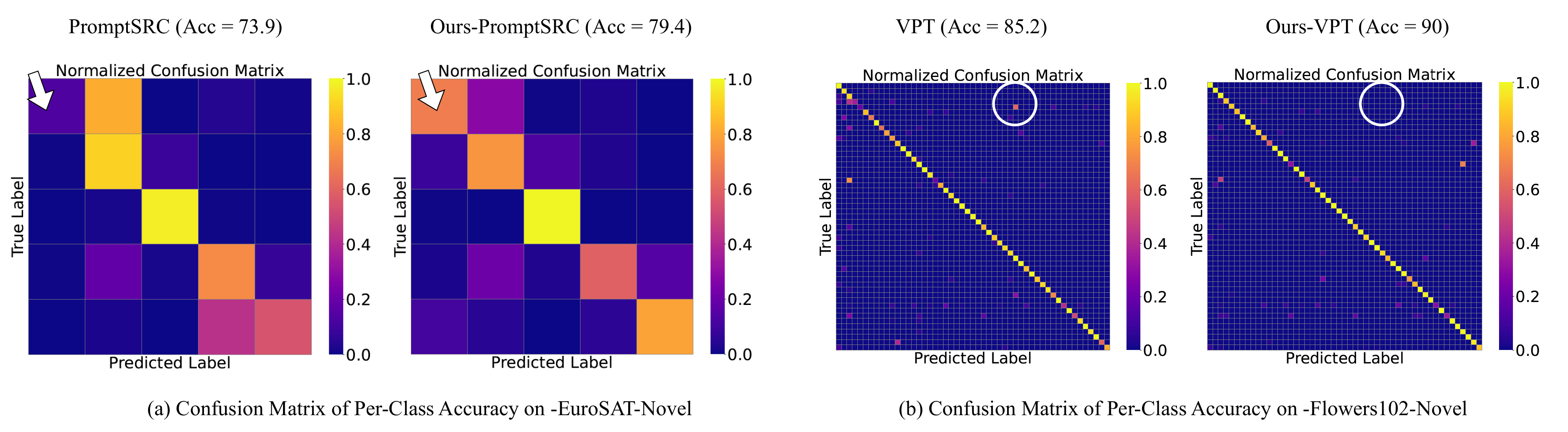}
    \caption{\textbf{The Confusion Matrix for Per-Class Accuracy.} For Figure (a), without our method, the category in the first row is misclassified as the second category. After using our method, the first category classification is successfully made to achieve the highest accuracy.For Figure (b), our method also significantly improves one misclassified subclass, thereby improving the overall accuracy on the entire task.}
    \label{fig:confusion}
\end{figure*}

\subsection{t-SNE Visualization}
We show more t-SNE visualization results in Figure \ref{fig:t-sne}. In Figure (a), it is evident that applying our $\mathcal{L}_{\text {Aligned}}$ method to LPT increases the distance between cluster centers of the green color point and the orange color points. This indicates that our method enhances the learned latent space, bringing it closer to real samples, strengthening the model's decision boundaries, and consequently improving its accuracy. Similar improvements are observed in Figures \ref{fig:t-sne} (b) (c) and (d). Additionally, the circle in Figure \ref{fig:t-sne} (a) and Figure (b) shows that by using our method, we separate the overlap clusters to no-overlap clusters, which also confirms the effectiveness of our $\mathcal{L}_{\text {Aligned}}$ method.

\subsection{Confusion Matrix Comparison} 
To conduct a more granular analysis of the performance improvements brought about by our method, we visualized the confusion matrices representing the accuracy for each category. The experimental results are illustrated in Figure \ref{fig:confusion}. In Figure \ref{fig:confusion} (a), in the PromptSRC classification experiment on the EuroSAT dataset, the highest value in the first row of the baseline confusion matrix deviated from the diagonal, representing the Pasture Land category, with an accuracy of only 13.2\%. Upon utilizing our $\mathcal{L}_{\text{Aligned}}$ loss function, the first row of the confusion matrix aligned with the diagonal, and the classification accuracy for Pasture Land improved to 68\%, which lead to the all-class accuracy improved to 79.4\%. Similarly in Figure \ref{fig:confusion} (b), the figure shows the classification experiments of VPT on the Oxford Flowers dataset. In the confusion matrix of the baseline model, we observed that the classification accuracy for the fifth category, English Marigold, deviated significantly from the diagonal, with an accuracy of only 20\%. After applying our proposed $\mathcal{L}_{\text{Aligned}}$ loss function, the classification accuracy for English Marigold increased to 90\%.

\section{Limitations}
Our method aims to construct relative representations in the latent space for cross-modal alignment between image and text modalities, utilizing both static and stochastic anchors. A significant limitation of this method is its high dependency on the selection of the Anchor. For instance, if the static anchor selected does not accurately capture the clustering characteristics of the targeted image category, it may result in biased cross-modal alignment, thereby adversely affecting the learning performance of the model. Additionally, in complex or non-standardized scenarios, finding a suitable static anchor point can be challenging, which constrains the general applicability of our approach

\section{Broader Impact}
\vspace{-0.1in}
Our proposed approach offers an effective technique applicable to visual language models characterized by an Image-Text dual-branch architecture, which is plug-and-play and can be integrated with many existing prompt tuning methods. Consequently, applying our method to the more sophisticated Prompt Tuning framework could yield further enhancements in performance. We leave these explorations for future research.

\end{document}